\documentclass{article} 
\usepackage{arxiv,times}

\usepackage{amsmath,amsfonts,bm}

\def\eqref#1{equation~\ref{#1}}

\def\1{\bm{1}}

\DeclareMathAlphabet{\mathsfit}{\encodingdefault}{\sfdefault}{m}{sl}
\SetMathAlphabet{\mathsfit}{bold}{\encodingdefault}{\sfdefault}{bx}{n}

\usepackage{comment}
\usepackage[utf8]{inputenc} 
\usepackage[T1]{fontenc}    
\usepackage{hyperref}       
\usepackage{url}            
\usepackage{booktabs}       
\usepackage{amsfonts}       
\usepackage{nicefrac}       
\usepackage{microtype}      
\usepackage{lipsum}
\usepackage[utf8]{inputenc}
\usepackage{fourier} 
\usepackage{array}
\usepackage{makecell}
\usepackage{graphicx}
\usepackage{subcaption}
\usepackage{color}
\usepackage{siunitx}
\usepackage{verbatim} 
\usepackage{textcomp}
\usepackage{framed}
\usepackage{amsmath} 
\usepackage{url}
\usepackage{pdfpages} 
\usepackage{pgfplots, pgfplotstable} 
\usepackage{algorithm}
\usepackage{algorithmic}
\usepackage{bm}
\usepackage{float} 
\pgfplotsset{compat=1.14}
\usepackage{wrapfig} 
\usepackage{appendix}

\definecolor{tb_color_1}{RGB}{245,124,0}
\definecolor{tb_color_2}{RGB}{0,167,247}
\definecolor{tb_color_3}{RGB}{235, 44, 108}
\definecolor{tb_color_4}{RGB}{1, 142, 125}
\definecolor{tb_color_5}{RGB}{153, 152, 153}

\pgfplotscreateplotcyclelist{tb}{
semithick,tb_color_1\\%
semithick,tb_color_2\\%
semithick,tb_color_3\\%
semithick,tb_color_4\\%
semithick,tb_color_5\\%
}

\newcommand{\AP}[1]{{\color{blue}{[}\textbf
{AP: #1}{]}}}
\newcommand{\KK}[1]{{\color{purple}{[}\textbf
{KK: #1}{]}}}

\newcommand{\NJG}[1]{{\color{orange}{[}\textbf
{NJG: #1}{]}}}
\newcommand{\TODO}[1]{{\color{red}{[}\textbf
{TODO: #1}{]}}}

\title{Campfire: 
Compressible, Regularization-Free, Structured Sparse Training for Hardware Accelerators} 

\author{Noah Gamboa, Kais Kudrolli, Anand Dhoot, and Ardavan Pedram \\
Department of Computer Science\\
Stanford University\\
Stanford, CA 94305, USA \\
\texttt{\{ngamboa, kudrolli, anandd, perdavan\}@stanford.edu} \\
}

\begin{document}

\maketitle

\vspace{-10pt}
\begin{abstract}
This paper studies structured sparse training of CNNs with a gradual pruning technique that leads to fixed, sparse weight matrices after a set number of epochs. 
We simplify the structure of the enforced sparsity so that it reduces overhead caused by regularization.
The proposed training methodology Campfire explores pruning at granularities within a convolutional kernel and filter.

We study various tradeoffs with respect to pruning duration, level of sparsity, and learning rate configuration.
We show that our method creates a sparse version of ResNet-50 and ResNet-50 v1.5 on full ImageNet while remaining within a negligible $<$1\% margin of accuracy loss.
To ensure that this type of sparse training does not harm the robustness of the network, we also demonstrate how the network behaves in the presence of adversarial attacks.  Our results show that with 70\% target sparsity, over 75\% top-1 accuracy is achievable.
\end{abstract}


\section{Introduction}
\label{sec:intro}

Pruning weights can compress a neural network into a smaller model that can fit into faster/smaller memory and therefore result in execution speedups \cite{EIE, han2015deep}. To increase the accuracy of sparse models, Han et al. \cite{han2015learning} and Mao et al. \cite{mao2017exploring} explore training the network dense after pruning.
The resulting network can maintain accuracy based on the specified level of sparsity~\cite{mostafa2019parameter, topruneornot, han2015deep}.

Structured sparsity, where a certain number of non-zeros is allowed across various cross-sections of the weight tensors, has been explored for RNNs and also CNNs. These methods aim to speed up computation and reach some final level of sparsity for deployment. Narang et al. \cite{narang2017exploring} have shown promising results for structured training of RNNs while sparse CNNs could not achieve the same performance \cite{mao2017exploring}.

Recent work has demonstrated that structurally-sparse training can speed up execution on GPUs~\cite{he2017channel, prunetrain, topruneornot}.
However, these training mechanisms add regularization (and thus computational overhead) to eliminate unnecessary weights. 
Regularization includes operations such as norm \cite{grouplasso}, involving division and square root, which are expensive for ASIC accelerators as they are atypical and high latency \cite{wu2018l1}.
While enforcing coarse-grain sparsity, PruneTrain \cite{prunetrain} provides significant speedups, but the final network contains a low degree of sparsity. Higher sparsity levels are necessary to offset the overheads incurred by indexing/compression/decompression \cite{kourtis2010data} and to fit models on memory-limited platforms such as mobile or embedded devices \cite{lin2019toward}.

Mostafa and Wang \cite{mostafa2019parameter} show that with adaptive sparse training and dynamic reallocation of non-zeros sparsity levels up to {80\%} can be achieved. However, to achieve an accuracy loss of 1.6\% an additional 10 epochs (100 total compared to the typical 90 epochs) of training are required. The main drawback is the overhead incurred while implementing such a technique on the target platform.
Continuous reconfiguration of the sparsity pattern is expensive in hardware as creating a compressed format is memory-intensive and energy-inefficient \cite{pedram2016dark}. Restructuring the sparsity pattern frequently requires recompression of data. The incurred memory accesses overshadow the savings when skipping computations with zeros and thus makes weight compression infeasible.

Our goal is to provide high levels of sparsity (>60\%) during training with minimal degradation of accuracy. Additionally, to make our training more memory-efficient and accelerator-friendly, we seek to make the sparsity structured, to remove irregular computations like regularization, and to avoid decompression/recompression at each training step. Our main motivating insight is that having a \textit{fixed} sparse multiply-accumulate pattern allows weight compression during training and can save compute and energy in hardware \cite{EIE}.

To achieve our goal, we introduce Campfire, a sparse training method that applies the techniques in Han et al. \cite{han2015learning} and Mao et al. \cite{mao2017exploring} at earlier stages in training within what we call the pruning era, usually a period of 20-30 epochs.
During the pruning era, we exploit one of the three proposed sparsity regimes, which have granularities of at most a whole kernel, to prune the network.
After this period, we fix the sparsity mask for the rest of the training. 
Since sparsification reduces the total number of computations \cite{zhu2019sparse, parashar2017scnn} and hardware accelerators have mechanisms to skip computations with zeros \cite{han2015learning}, fixing the mask early in training results in speedups and thus makes an earlier shorter pruning era ideal.
We seek to find this ideal pruning era by characterizing different combinations of pruning era length and start epoch with our original goal of high sparsity and high accuracy in mind.

As such, we explore the impact of various pruning granularities, sparsity levels, and learning-rate schedules on the network's convergence as well as adversarial robustness for CNNs like ResNet-50~\cite{resnet50} on ImageNet and Tiny Imagenet~\cite{tinyimagenet}.




Recent literature~\cite{wang2018adv} has shown that adversarial attacks are more successful on pruned neural networks than they are on regular neural networks. 
Given the danger of adversarial attacks in real world situations, we find that it is important to evaluate our sparsity techniques under adversarial robustness. 
We leverage the FGSM mechanism~\cite{goodfellow2014explaining} to evaluate the adversarial robustness on our sparse models.
This paper makes the following contributions:
\begin{enumerate}
     \item  We propose a mechanism to train and prune a convolutional network during the earlier stages of training such that this sparsity can be harvested for the computational speedups. To do this, we fix the sparse weight masks for the remainder of the training. 
     \item For fully connected sparsification, we eliminate blocks of fully connected weights based on their connection to the zeros in the previous convolutional layer.
      \item We enforce structural, regularization free, magnitude-based pruning across two distinct dimensions and a combined version. These dimensions are inside convolution window $R \times S$ and across input/output feature matrix ($CK$).
    \item Our sparse models are as robust to adversarial FGSM attacks as fully dense models. 
    \item We demonstrate that early stage dense training is crucial for maintaining high accuracy.
    
    \item The proposed technique is tolerant to sparsity levels of up to 60-70\% with under 1\% accuracy degradation. We can compensate by scheduling an extra learning rate drop and training for an extra 10 epochs.

\end{enumerate}

The rest of the paper is organized as follows. Section~\ref{sec:methodology} explains our pruning methodology. Section~\ref{sec:setup} describes the experimental setup framework. Section~\ref{sec:results} presents results and discusses their interpretation. Section~\ref{sec:related} presents the related work. Section~\ref{sec:conclusion} concludes the paper.


\section{Pruning Methodology}
\label{sec:methodology}

\begin{figure}[t]
   \vspace{0pt}
   \centering
        \begin{subfigure}[b]{0.22\textwidth}
           \includegraphics[width=\textwidth]{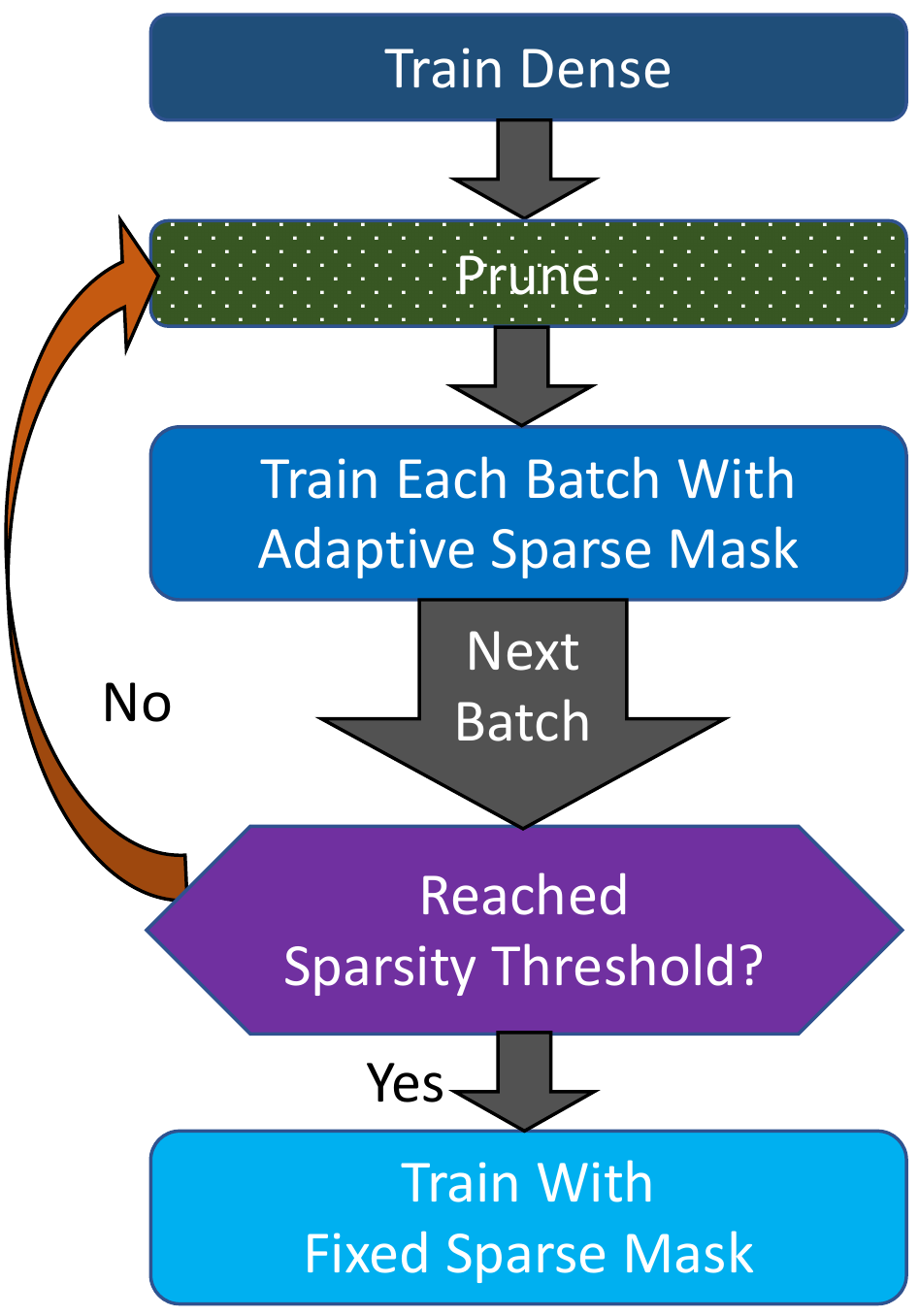}
           \caption{}
           \label{fig:chart}
        \end{subfigure}
        \quad
        \begin{subfigure}[b]{0.75\textwidth}
           \includegraphics[width=\textwidth]{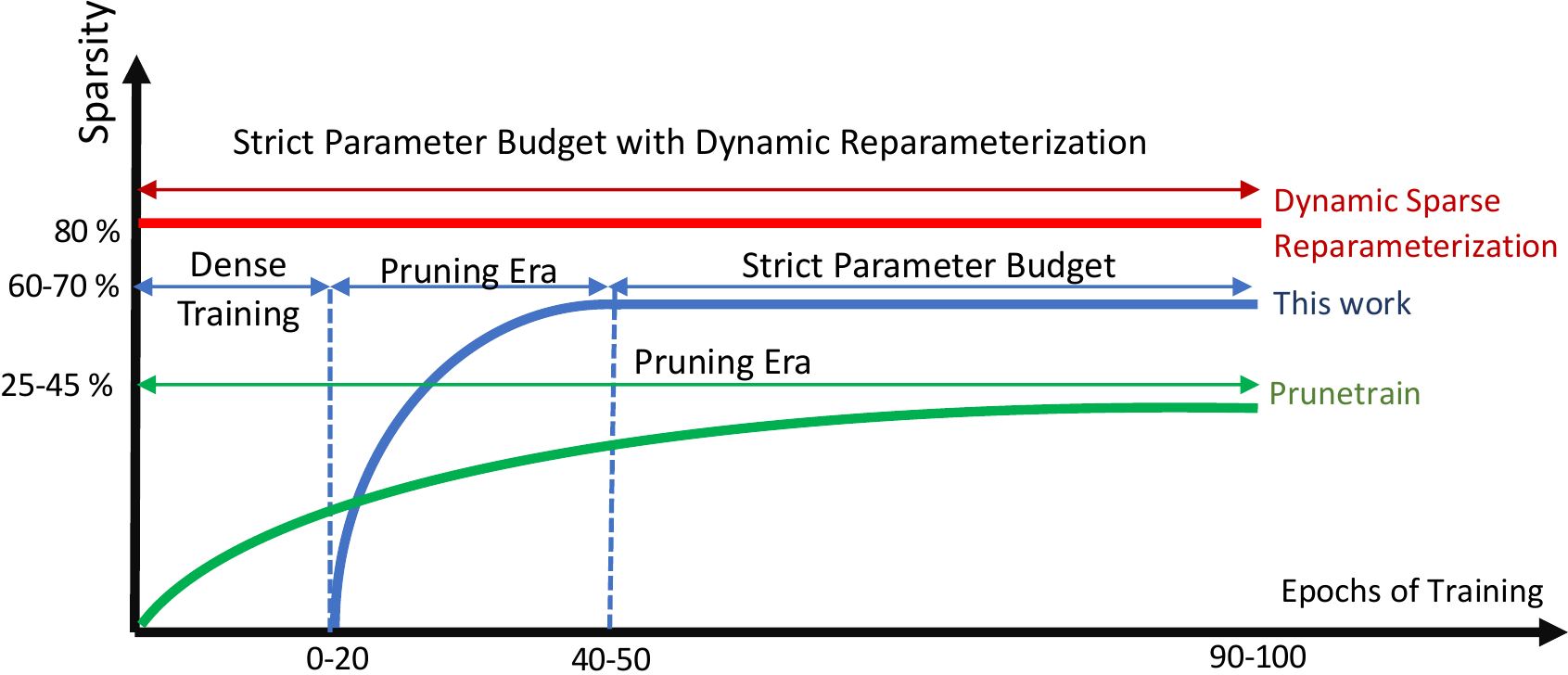}
          \caption{}
           \label{fig:timing}
        \end{subfigure}

\vspace{-10pt}
\caption{(a) Our general sparsity mechanism in which we update the sparsity mask after each batch until we reach a desired level of sparsity. (b) Schedule comparison between this work, Mostafa ang Wang \cite{mostafa2019parameter}, and PruneTrain \cite{prunetrain}. Our work has the shortest pruning era and more gradually reaches our final sparsity. }
\label{fig:pruning_methodology}
\end{figure}

Our proposed pruning mechanism works by always pruning the weights of smallest magnitude after each weight update. After a forward and backward pass (one batch update), the model is pruned. If a weight is already zero, the gradient is also set to zero. This means that once a weight becomes zero, it will remain zero for the rest of the training period. 

This mechanism is similar to Han et al. \cite{han2015learning}, except that we only prune in the earlier stages of the training as opposed to post training. 
Additionally, this work is similar to Narang et al. \cite{narang2017exploring} although we set the sparsity threshold instead of using a heuristic to calculate it.
We chose this pruning mechanism because of its negligible computational overhead.

In our pruning algorithm, 
the sparsity threshold refers to the percentage of weights in the network that are currently pruned. Before or during the first epoch of pruning, we will have a sparsity threshold of zero. 
As we continue training, we gradually increase the sparsity threshold so that by the final epoch of pruning the network sparsity will have reached our final, desired threshold.
This gradual increase is achieved by setting the threshold as shown in the following equation in each epoch within the pruning window:
\begin{align}
    \text{sparsity\_threshold} = s_f - (s_i + s_f) (1 - \frac{e_c - e_i}{l_p})^r
\end{align}
Where $s_f$ is the final desired sparsity, $s_i$ is the initial sparsity (always 0 in our case), $e_c$ is the current epoch, $e_i$ is the initial epoch of pruning, $l_p$ is the length of the \textit{pruning era}, and $r$ controls how fast or slow the threshold increases exponentially  (in our case $r=3$).

We define the \textit{pruning era} to be the epochs between the first and final epochs of pruning depicted in Figure~\ref{fig:timing}.

Finally, we evaluate the pruning mask after every training step until we reach the final epoch of pruning. After the final epoch, the pruned values in the network will remain zero for the rest of training; no new pruning will occur, and only the non-zero weights will be updated. 

\begin{figure}[t]
   \vspace{0pt}
   \centering
        \begin{subfigure}[b]{0.31\textwidth}
           \includegraphics[width=\textwidth]{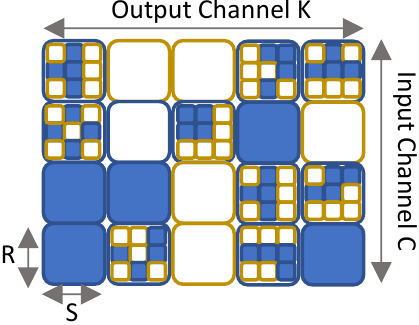}
           \caption{}
           \label{fig:window_pruning}
        \end{subfigure}
        \quad
        \begin{subfigure}[b]{0.3\textwidth}
           \includegraphics[width=\textwidth]{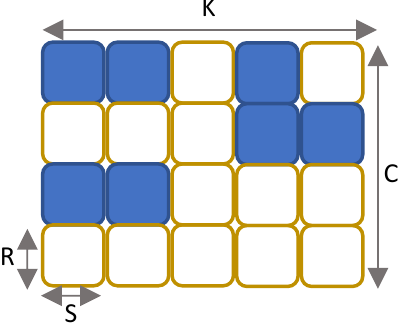}
          \caption{}
           \label{fig:ck_pruning}
        \end{subfigure}
        \quad
        \begin{subfigure}[b]{0.15\textwidth}
          \includegraphics{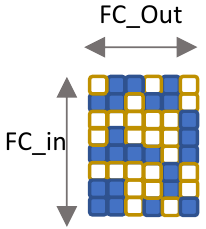}
          \caption{}
           \label{fig:fc_pruning}
        \end{subfigure}
        \quad
        \begin{subfigure}[b]{0.15\textwidth}
          \includegraphics{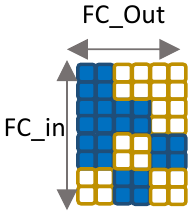}
          \caption{}
           \label{fig:block_fc_pruning}
        \end{subfigure}
 \vspace{-10pt}
 \caption{
This figure illustrates sparsity across different weight dimensions and different granularities. (a) \textit{Window sparsity}, which prunes the 5 smallest weights in a 3$\times$3 window. (b)\textit{CK pruning} where whole R$\times$S convolutional kernels are pruned. (c) Sparsity in fully-connected weights. (d) Block sparsity in a fully-connected layer.}

\label{fig:3x3convprune}
\vspace{0pt}
\end{figure}

\subsection{Pruning Methodology by Layer}
\label{sec:layer-pruning}

Pruning the smallest magnitude weights in the \textit{entire} network is inefficient because it involves sorting the weights over the network. Instead, we prune the smallest magnitude weights or sum of weights, within a certain locale of the network. When pruning, we examine each layer individually and apply a separate technique to evaluate which weights to prune, depending on the type of layer we are currently pruning. 

\subsubsection{Convolutional Layer Pruning}

\paragraph{Window pruning for 3x3 Convolutional Layers}
Figure~\ref{fig:window_pruning} shows the result of a pruned 3$\times$3 convolutional weight tensor under the window pruning method. 
In this scheme, window layer pruning refers to pruning of weights within the 3$\times$3 convolution kernels. 
We allow a maximum number of non-zero values for each kernel in the 3$\times$3 convolutional layers and eliminate the weights of smallest magnitude. We set this fixed value so that a hardware accelerator could allocate a fixed dataflow based on these maximum values \cite{EIE}.

\paragraph{CK Pruning Methodology}

\begin{algorithm}
\footnotesize
\caption{CK Pruning Algorithm}
\label{alg:ck_pruning}
\begin{algorithmic}
\STATE generate\_ck\_sparsity\_mask($\theta_{layer}$, sparsity\_threshold):
    \FOR{$\theta$ in $\theta_{layer}$}
        \FOR{all c in C} 
            \FOR{all k in K}
                \STATE kernel\_max$_{c,k}$ = max($\theta_{c,k}$)
            \ENDFOR
            \STATE cutoff\_index = size($\theta_c$) $*$ sparsity\_threshold
            \STATE n = max(cutoff\_index, size($\theta_c$) $-$
            max\_non\_zero $-$ 1)
            \STATE cutoff\_value = $n^{th}$ largest value in kernel\_max$_c$
            \FOR{all k in K}
                \STATE mask$_{c,k}$ $=$ 1 if kernel\_max$_{c,k}$ $>$ cutoff\_value, else 0
            \ENDFOR
        \ENDFOR
    \ENDFOR
\end{algorithmic}
\end{algorithm}

Figure~\ref{fig:ck_pruning} shows the result of a pruned 3$\times$3 convolutional weight tensor under the CK pruning method. 
In this scheme, the weights of a certain layer can be viewed as a CK matrix of R$\times$S kernels. 
The CK pruning method involves pruning the 3$\times$3 convolutions along the channel and kernel dimensions of each convolutional filter, i.e., we prune whole kernels (CK matrix of R$\times$S windows) at once and can ultimately prune all the input channels in an output channel. 
As defined by Algorithm~\ref{alg:ck_pruning}, we determine which filter to prune by examining the max of the magnitudes of all the weights in a kernel, which is the max of nine weights. 
This max is used to evaluate whether the whole kernel should be pruned or not.

\paragraph{Combined Pruning Methodology}
To combine window and CK pruning, we introduce a \textit{combined pruning method}.
As shown by appendix Algorithm~\ref{alg:combined_pruning} in the Appendix, in a given epoch we first apply window pruning to each 3$\times$3 convolutional layer at a fraction of the \textit{sparsity threshold} for that epoch. Then, we prune the remaining fraction of the \textit{sparsity threshold} with CK Pruning. Combined pruning has a \textit{window pruning threshold} hyperparameter (between 0 and 1) that determines how much window pruning is done. It is multiplied with the current epoch's sparsity threshold to get a new threshold used for the window pruning phase. We set this parameter to 0.8 in our experiments.

\begin{figure}[t]
    \centering
    \vspace{-10pt}
    \includegraphics[width=\textwidth]{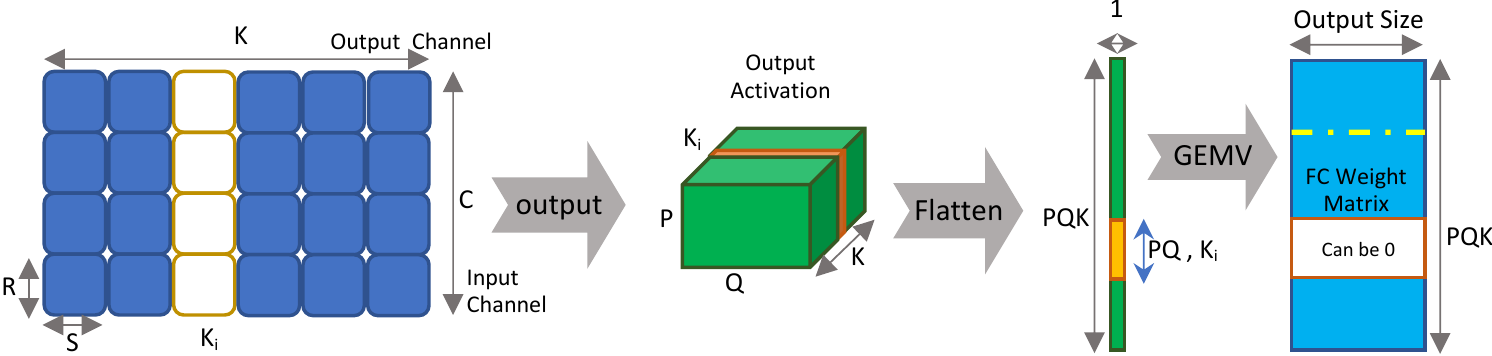}
    \vspace{-15pt}
    \caption{FC pruning taking advantage of the sparsity in the previous convolution layer.}
    \vspace{-5pt}
    \label{fig:FC_pruning}
\end{figure}
 
\subsubsection{Fully Connected Pruning}
Like pruning for convolutional layers, we apply a two-tier pruning scheme from Mao et al. \cite{mao2017exploring} for fully connected layers: micro-level pruning within a block and macro-level pruning that eliminates entire blocks. 

\paragraph{Block  FC Pruning} Figure~\ref{fig:block_fc_pruning} refers to pruning of individual blocks. Here, we prune an entire n$\times$n (n$<$5) window within the dense layer and create coarse grained sparsity. To do this, we sum the magnitude of the weights in each window and prune the windows with the smallest magnitude. 

\paragraph{Fine FC Pruning} Figure~\ref{fig:fc_pruning} refers to the pruning of individual weights. Here, we prune the individual weights in the entire FC Layer, where we compare the magnitude of all the weights to each other. 

The produced zero patterns in the last convolution layer allow for eliminating more weights in fully connected layer as depicted in Figure~\ref{fig:FC_pruning}.  
If all the $C$ windows for a specific $K_i$ are zeros, the output activation for the corresponding $K_i$ is also zero. 
The corresponding neurons in the following fully connected layer are therefore receiving zero input activations and can be eliminated along with their associated weights. 
This enables us to get sparsity without having to evaluate the weights in the fully connected layer. 

When pruning just the small weights in the FC layer, one can inadvertently cut off relevant connections between the input and output layers. Accordingly, we structure the pruning mechanism such that each output neuron should be influenced by the input. This means every column in the weight matrix of the fully connected layer in Figure~\ref{fig:FC_pruning} has at least one non-zero element.


    
\section{Experimental Setup}
\label{sec:setup}

To validate each type of pruning (window, CK, or combined) we selected ResNet-50 \cite{resnet50} v1 and v1.5 with the ImageNet and/or Tiny-ImageNet \cite{tinyimagenet} datasets. We evaluated each pruning method by varying \textit{sparsity levels} and \textit{pruning era}. Smaller batch (64) experiments were each run on one NVIDIA RTX 2080 GPU, and larger batch (256) experiments were run on 4. Each network was implemented in PyTorch \footnote{\url{https://pytorch.org/}}.

We experimented with ResNet-50 v1.5, in addition to v1, to explore how changing the network structure would affect the top-1 accuracy. For window pruning, we tested with ResNet-50 v1 on Tiny-ImageNet as well as ResNet50 v1 and v1.5 on ImageNet to compare the impact of strided convolutions on our sparse training.
Also, we experimented with the learning rate schedule of the training regime. Our typical  schedule for ResNet-50 v1.5 included learning rate drops at epochs 30, 60, and 90, but we experimented with placing the last drop at epoch 80 instead. 
For the majority of our experiments, we used batch size 64 as this is what could fit in one of our GPUs. As suggested by Krizhevsky \cite{krizhevsky2014weird}, we scaled the starting learning rate by $\frac{1}{\sqrt{4}} = \frac{1}{2}$ to 0.05 in order to compensate for the smaller batch size. We also showed results with batch size 256 spread across 4 GPUs and starting the learning rate at 0.1.

\subsection{Sparse Training Experiments}

Tiny-Imagenet, which has 100,000 training images~\cite{tinyimagenet}, is an easier task than full ImageNet (1.2 million images~\cite{imagenet}), but it takes less time to train and is still somewhat predictive of performance on ImageNet as it contains the same type of images. Accordingly, we ran with Tiny-Imagenet on ResNet-50 v1 as a preliminary test of our pruning methods. As dense training with this benchmark only requires 40 epochs to converge and we wanted to fix the sparsity mask as early in the training as possible, we used a pruning era of epochs 0-10. Early experiments showed this to be sufficient to achieve full accuracy with Window and CK pruning.

In order to find the ideal pruning era and compare the relative performance of our pruning methods (Window, CK, and Combined), we mainly used ResNet-50 v1 and ResNet-50 v1.5 with ImageNet. As trying every permutation of pruning era length and starting epoch was infeasible for us resource-wise, we used our early experiments to guide our search. Initially, with window and CK pruning we experimented with pruning at the beginning of training (epochs 0-30). While these experiments were promising with Window pruning, CK pruning was not effective early in training. Furthermore, we tried a shorter \textit{pruning era} (0-20) with Window but found this caused a large drop in accuracy, so we mostly moved away from pruning in the early epochs. 

Next, we hypothesized that epoch 30 would be a suitable epoch to stop pruning as this is the epoch of the first learning rate decrease. Results with all methods using a pruning era of 30-50 were promising, so we searched around starting epoch 30. Accordingly, we adopted a similar approach to Han et al. \cite{han2015learning} to train with all pruning methods by setting the \textit{first epoch of pruning} to 20, 30, or 40 and the \textit{pruning era} to 20 or 30 epochs. In these combinations, we experimented with the following sparsities (and report the most important results): 20, 40, 60, 70, 80, and 90\%.

We did experiment with earlier starting epochs (10, 15) and a shorter pruning era lengths (10); however, these caused large losses in accuracy, so we did not pursue them fully. We did not pursue later starting epochs or longer pruning eras because our goal was to fix the sparsity mask early and prune for as few epochs as possible.




In each of network/dataset combinations, we compare to a densely trained baseline and other works of literature that performed corresponding experiments.

\subsection{Adversarial Robustness}

Since there was evidence that increasing sparsity lowers adversarial robustness \cite{wang2018adv}, we evaluated this robustness in our models.
To do so, we applied \textit{Fast Gradient Sign Method} (FGSM) attacks \cite{goodfellow2014explaining} on one of our sparse models, to generate its own adversarial examples, and measured the validation accuracy again. 
We used the same validation set as ImageNet and applied the attack's image transformation to each input image.
Moreover, we experimented with a variety of different \(\epsilon\) in order to see how our accuracy decayed. 
Lastly, in our experiments we leveraged the examples provided in Pytorch tutorials. \footnote{\url{https://pytorch.org/tutorials/beginner/fgsm_tutorial.html}}

\section{Results}
\label{sec:results}



\subsection{ResNet-50 on Tiny-Imagenet}

From our experiments  with Tiny-Imagenet, we see that even with up to 80\% sparsity, both window and CK pruning are able to achieve levels of accuracy comparable to the dense baseline. CK pruning performs even better than the baseline. Our results are shown in Table~\ref{tab:window_ck_resnet50_timagenet} below.

\begin{table}[t]
    \vspace{-5pt}
    \centering
    \footnotesize
    \begin{tabular}{|c|c|c|c|c|c|c|} \hline
         & \multicolumn{3}{c|}{\thead{Window}} & \multicolumn{3}{c|}{\thead{CK}} \\ \hline
         \thead{Model \\ Sparsity [\%]}& \thead{Accuracy [\%]} & \thead{Epoch of \\Convergence} & \thead{True \\ sparsity} & \thead{Accuracy [\%]} & \thead{Epoch of \\Convergence} & \thead{True sparsity [\%]}\\ \hline
         0 & 52.03 & 40 & 0.01 & 51.84 & 31 & 0.021\\
         20     & 50.97 & 40 & 0.48 & 52.40 & 31 & 0.207 \\
         40     & 51.62 & 39 & 0.58 & 51.79 & 31 & 0.404 \\
         60     & 52.09 & 40 & 0.69 & 52.56 & 31 & 0.602 \\
         80     & 51.16 & 36 & 0.84 & 52.07 & 31 & 0.800 \\ \hline
    \end{tabular}
    \vspace{0pt}
    \caption{Best accuracy for different sparsity levels for ResNet-50 on Tiny-Imagenet with Window and CK Pruning. Both methods see high accuracy close to, or even above the baseline in the cases of Window at 60\%, CK at 20\%, 60\% and 80\%. CK converges to the final epoch at the same epoch as the baseline while window converges at the same epoch or earlier in the cases of 40\% and 60\%.}
    \label{tab:window_ck_resnet50_timagenet}
    \vspace{-5pt}
\end{table}

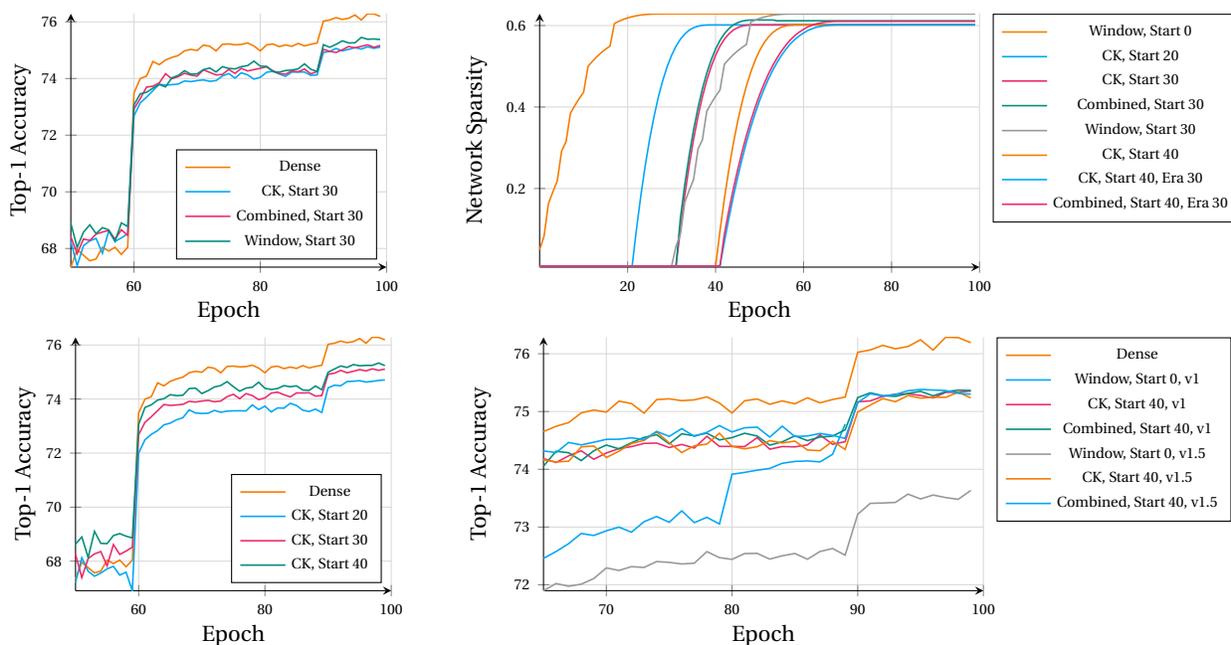
\begin{figure}[t]
    \vspace{0pt}
    \centering
    \begin{tikzpicture}
    \begin{axis}[cycle list name=tb, 
                    width=0.35\textwidth,
                    height=0.30\textwidth,
                    grid=both,
                    grid style={solid,gray!30!white},
                    axis lines=middle,
                    xlabel={Epoch},
                    ylabel={Top-1 Accuracy},
                    x label style={at={(axis description cs:0.5,-0.1)},anchor=north,font=\footnotesize},
                    y label style={at={(axis description cs:-0.1,.5)},rotate=90,anchor=south,font=\footnotesize},
                    xmax=100,
                    xmin=50,
                    legend pos=south east,
                    tick label style={font=\tiny},
                    legend style={font=\tiny}
                    ]
    \addplot table [x=, y=resnet1_5_50_baseline-dense-lr0-05out.out, col sep=comma]
    {ck_imagenet_resnet50v1_5/sparse60/dense/Val_Prec_1_Avg.csv};
    \addlegendentry{Dense}
    \addplot table [x=, y=resnet1_5_50_baseline-epoch-30-cksparse60-lr0-05out.out, col sep=comma]
    {ck_imagenet_resnet50v1_5/sparse60/start30/Val_Prec_1_Avg.csv};
    \addlegendentry{CK, Start 30}
    \addplot table [x=, y=resnet1_5_50_baseline-epoch-30-cksparse60-lr0-05-intra-out.out, col sep=comma]
    {ck_imagenet_resnet50v1_5/sparse60/start30/Val_Prec_1_Avg.csv};
    \addlegendentry{Combined, Start 30}
    \addplot table [x=, y=resnet1_5_50_baseline-epoch-30-window60-lr0-05out.out, col sep=comma]
    {ck_imagenet_resnet50v1_5/sparse60/start30/Val_Prec_1_Avg.csv};
    \addlegendentry{Window, Start 30}

    \end{axis}
    \end{tikzpicture}
\qquad
     \begin{tikzpicture}
    \begin{axis}[cycle list name=tb, 
                    width=0.45\textwidth,
                    height=0.3\textwidth,
                    grid=both,
                    grid style={solid,gray!30!white},
                    axis lines=middle,
                    xlabel={Epoch},
                    ylabel={Network Sparsity},
                    x label style={at={(axis description cs:0.5,-0.1)},anchor=north,font=\footnotesize},
                    y label style={at={(axis description cs:-0.1,.5)},rotate=90,anchor=south,font=\footnotesize},
                    xmax=100,
                    xmin=0,
                    legend pos=outer north east,
                    tick label style={font=\tiny},
                    legend style={font=\tiny}
                    ]
    \addplot table [x=, y=resnet1_5_50_baseline-epoch-0-window60-lr0-05out.out, col sep=comma]
    {ck_imagenet_resnet50v1_5/sparse60/start0/Val_Sparsity.csv};
    \addlegendentry{Window, Start 0}
    \addplot table [x=, y=resnet1_5_50_baseline-epoch-20-cksparse60-lr0-05out.out, col sep=comma]
    {ck_imagenet_resnet50v1_5/sparse60/start20/Val_Sparsity.csv};
    \addlegendentry{CK, Start 20}
    \addplot table [x=, y=resnet1_5_50_baseline-epoch-30-cksparse60-lr0-05out.out, col sep=comma]
    {ck_imagenet_resnet50v1_5/sparse60/start30/Val_Sparsity.csv};
    \addlegendentry{CK, Start 30}
    \addplot table [x=, y=resnet1_5_50_baseline-epoch-30-cksparse60-lr0-05-intra-out.out, col sep=comma]
    {ck_imagenet_resnet50v1_5/sparse60/start30/Val_Sparsity.csv};
    \addlegendentry{Combined, Start 30}
    \addplot table [x=, y=resnet1_5_50_baseline-epoch-30-window60-lr0-05out.out, col sep=comma]
    {ck_imagenet_resnet50v1_5/sparse60/start30/Val_Sparsity.csv};
    \addlegendentry{Window, Start 30}
    \addplot table [x=, y=resnet1_5_50_baseline-epoch-40-cksparse60-lr0-05out.out, col sep=comma]
    {ck_imagenet_resnet50v1_5/sparse60/start40/Val_Sparsity.csv};
    \addlegendentry{CK, Start 40}
    \addplot table [x=, y=resnet1_5_50_baseline-era30-epoch-40-cksparse60-lr0-05out.out, col sep=comma]
    {ck_imagenet_resnet50v1_5/sparse60/start40_era30/Val_Sparsity.csv};
    \addlegendentry{CK, Start 40, Era 30}
    \addplot table [x=, y=resnet1_5_50_baseline-era30-lr05-epoch-40-intra-combo-cksparse60-out.out, col sep=comma]
    {ck_imagenet_resnet50v1_5/sparse60/start40_era30/Val_Sparsity.csv};
    \addlegendentry{Combined, Start 40, Era 30}

    \end{axis}
    \end{tikzpicture}

    \begin{tikzpicture}
    \begin{axis}[cycle list name=tb, 
                    width=0.35\textwidth,
                    height=0.30\textwidth,
                    grid=both,
                    grid style={solid,gray!30!white},
                    axis lines=middle,
                    xlabel={Epoch},
                    ylabel={Top-1 Accuracy},
                    x label style={at={(axis description cs:0.5,-0.1)},anchor=north,font=\footnotesize},
                    y label style={at={(axis description cs:-0.1,.5)},rotate=90,anchor=south,font=\footnotesize},
                    xmax=100,
                    xmin=50,
                    legend pos=south east,
                    tick label style={font=\tiny},
                    legend style={font=\tiny}
                    ]
    \addplot table [x=, y=resnet1_5_50_baseline-dense-lr0-05out.out, col sep=comma]
    {ck_imagenet_resnet50v1_5/sparse60/dense/Val_Prec_1_Avg.csv};
    \addlegendentry{Dense}
    \addplot table [x=, y=resnet1_5_50_baseline-epoch-20-cksparse60-lr0-05out.out, col sep=comma]
    {ck_imagenet_resnet50v1_5/sparse60/start20/Val_Prec_1_Avg.csv};
    \addlegendentry{CK, Start 20}
    \addplot table [x=, y=resnet1_5_50_baseline-epoch-30-cksparse60-lr0-05out.out, col sep=comma]
    {ck_imagenet_resnet50v1_5/sparse60/start30/Val_Prec_1_Avg.csv};
    \addlegendentry{CK, Start 30}
    \addplot table [x=, y=resnet1_5_50_baseline-epoch-40-cksparse60-lr0-05out.out, col sep=comma]
    {ck_imagenet_resnet50v1_5/sparse60/start40/Val_Prec_1_Avg.csv};
    \addlegendentry{CK, Start 40}
    
    \end{axis}
    \end{tikzpicture}
\qquad 
    \begin{tikzpicture}
    \begin{axis}[cycle list name=tb, 
                    width=0.45\textwidth,
                    height=0.30\textwidth,
                    grid=both,
                    grid style={solid,gray!30!white},
                    axis lines=middle,
                    xlabel={Epoch},
                    ylabel={Top-1 Accuracy},
                    x label style={at={(axis description cs:0.5,-0.1)},anchor=north,font=\footnotesize},
                    y label style={at={(axis description cs:-0.1,.5)},rotate=90,anchor=south,font=\footnotesize},
                    xmin=65,
                    xmax=100,
                    legend pos=outer north east,
                    tick label style={font=\tiny},
                    legend style={font=\tiny}
                    ]

    \addplot table [x=, y=resnet1_5_50_baseline-dense-lr0-05out.out, col sep=comma]
    {ck_imagenet_resnet50v1_5/sparse60/dense/Val_Prec_1_Avg.csv};
    \addlegendentry{Dense}
    \addplot table [x=, y=maxnonzero_4_60_0_30_nofc_mlplr_steps, col sep=comma]
    {imagenet_resnet50_window_sparsity/Val_Prec_1_Avg.csv};
    \addlegendentry{Window, Start 0, v1}
    \addplot table [x=, y=resnet1_50_baseline-epoch-40-cksparse60-lr0-05out.out, col sep=comma]
    {ck_imagenet_resnet50v1_5/v1/Val_Prec_1_Avg.csv};
    \addlegendentry{CK, Start 40, v1}
    \addplot table [x=, y=resnet1_50_baseline-lr05-epoch-40-intra-combo-cksparse60-out.out, col sep=comma]
    {ck_imagenet_resnet50v1_5/v1/Val_Prec_1_Avg.csv};
    \addlegendentry{Combined, Start 40, v1}
    
    \addplot table [x=, y=resnet1_5_50_baseline-epoch-0-window60-lr0-05out.out, col sep=comma]
    {ck_imagenet_resnet50v1_5/sparse60/start0/Val_Prec_1_Avg.csv};
    \addlegendentry{Window, Start 0, v1.5}
    \addplot table [x=, y=resnet1_5_50_baseline-epoch-40-cksparse60-lr0-05out.out, col sep=comma]
    {ck_imagenet_resnet50v1_5/sparse60/start40/Val_Prec_1_Avg.csv};
    \addlegendentry{CK, Start 40, v1.5}
    \addplot table [x=, y=resnet1_5_50_baseline-lr05-epoch-40-intra-combo-cksparse60-out.out, col sep=comma]
    {ck_imagenet_resnet50v1_5/sparse60/start40/Val_Prec_1_Avg.csv};
    \addlegendentry{Combined, Start 40, v1.5}
    \end{axis}
    \end{tikzpicture}
\vspace{-10pt}
\caption{top left: Convergence plots of all pruning methods with first epoch of pruning 30,
top right: Sparsity plot of all methods,
bottom left: and CK starting pruning at different epochs,
bottom right: ResNet v1 and v1.5 at 60\% sparsity for 90 epochs.}
\vspace{-10pt}
\label{fig:sparse60_resnet50v15_imagenet}
\end{figure}


\subsection{ResNet-50 on Imagenet}

Our ResNet-50 v1.5 experiments (Table~\ref{tab:main_results100} and Appendix Figure~\ref{fig:start30_resnet50v15_imagenet}) with the  \textit{first epoch of pruning} at epoch 30 show that all of our pruning methods are able to achieve over 73\% accuracy, and we can achieve above 74\% accuracy up to 70\% sparsity.  

By comparing the sparsity curves of the window, CK, and combined pruning runs in Figure~\ref{fig:sparse60_resnet50v15_imagenet} (top right), we observe that the sparsity of window pruning is not as smooth as the other methods. This is likely indicative of the more rigid structure of CK and combined pruning, which causes the degree of sparsity to be much more uniform from epoch to epoch. 
Figure~\ref{fig:sparse60_resnet50v15_imagenet} (top left, bottom right) also shows on ResNet v1.5, the window is slightly better than the CK and combined, which have similar performance, but the window is worse than the other two on ResNet v1. 
Furthermore, starting the \textit{pruning era} later improves performance (Figure~\ref{fig:sparse60_resnet50v15_imagenet}-(bottom left)).

Table~\ref{tab:main_resnet_v1} shows that on ResNet-50 v1, our methods can achieve between 0.1-0.3\% less than the baseline. 
Here, we do not compare to compression focused methods as they take around 180 epochs of training if aiming for levels of accuracy that reported. If not, they have much worse accuracy numbers without providing structured sparsity and without the potential of computation savings during training.

\begin{table}[t]
    \vspace{0pt}
    \centering
    \begin{tabular}{|c|c|c|c|c|} \hline
    Model Sparsity (\%) & 40 & 60 & 70 & 80 \\ \hline \hline
    Dense (76.29) & - & - & - & - \\ \hline
    CK, start 40 & 75.82 (-0.46) & 75.33 (-0.95) & 74.92 (-1.36) & \textbf{74.16 (-2.12)} \\
    CK, start 30 & 75.84 (-0.45) & 75.12 (-1.17) & 74.72 (-1.56) & 73.66 (-2.63) \\
    CK, start 20 & 75.55 (-0.74) & 74.71 (-1.58) & 74.32 (-1.96) & 72.79 (-3.50) \\ \hline
    Combined, start 40 & \textbf{75.89 (-0.39)} & \textbf{75.38 (-0.90)} & \textbf{75.07 (-1.21)} & 74.02 (-2.27)\\ 
    Combined, start 30 & 75.84 (-0.45) & 75.16 (-1.12) & 74.48 (-1.80) & 73.66 (-2.63) \\ 
    Combined, start 20 & 75.72 (-0.57) & 74.75 (-1.53) & 74.26 (-2.02) & 72.97 (-3.32) \\ \hline
    Window, start 0 & - & 73.63 (-2.65) & 72.79 (-3.50) & 70.25 (-6.04) \\ 
    Window, start 30 & - & 75.45 (-0.84) & 74.65 (-1.63) & 73.31 (-2.98) \\ \hline
    CK, start 40, era 40-70    & - & 75.52 (-0.77) & \textbf{75.16 (-1.13)} & - \\
    Combined, start 40, era 40-70 & - & \textbf{75.56 (-0.73)} & 75.14 (-1.15) & - \\ \hline
    \end{tabular}
    \vspace{5pt}
    \caption{Main results, top-1 accuracy on ResNet-50 v$1.5$ after 100 epochs.}
    \vspace{-10pt}
    \label{tab:main_results100}
\end{table}

\begin{table}[t]
    \centering
    \begin{tabular}{|c|c|c|c|c|} \hline
    Model Sparsity (\%) & 40 & 60 & 70 & 80 \\ \hline \hline
    Dense (75.25) & - & - & - & - \\ \hline
    CK, start 40 & 74.93 (-0.32) & 74.6 (-0.65) & 74.21 (-1.04) & \textbf{73.34 (-1.91)} \\
    CK, start 30 & \textbf{74.96 (-0.29)} & 74.25 (-1.00) & 73.83 (-1.42) & - \\
    CK, start 20 & 74.74 (-0.51) & 73.84 (-1.41) & 73.37 (-1.88) & 72.49 (-2.76) \\ \hline
    Combined, start 40 & 74.91 (-0.34) & \textbf{74.75 (-0.50)} & \textbf{74.36 (0.89)} & 73.22 (-2.03) \\ 
    Combined, start 30 & 74.94 (-0.31) & 74.41 (-0.84) & 73.87 (-1.38) & - \\ 
    Combined, start 20 & 74.73 (-0.52) & 73.65 (-1.60) & - & 72.22 (-3.03) \\ \hline
    Window, start 0 & - & 72.63 (-2.62) & 71.79 (-3.46) & 69.48 (-5.77) \\ 
    Window, start 30 & - & 74.28 (-0.97) & 73.88 (-1.37) & 72.63 (-2.62) \\ \hline
    CK, start 40, era 40-70    & - & 74.67 (-0.58) & 74.34 (-0.91) & - \\
    Combined, start 40, era 40-70 & - & \textbf{74.77 (-0.48)} & \textbf{74.38 (-0.87)} & - \\ \hline
    \end{tabular}
    \vspace{5pt}
    \caption{Main results, top-1 accuracy on ResNet-50 v$1.5$ after 90 epochs.}
    \label{tab:main_results90}
\end{table}

\begin{table}[!t]
    \footnotesize
    \vspace{-10pt}
    \centering
    \begin{tabular}{|c||c|c||c|c||} \hline
    & \multicolumn{2}{c||}{Epoch 90} & \multicolumn{2}{c||}{Epoch 100} \\ \hline
    Model & Sparsity [\%] & Accuracy [\%] & Sparsity [\%] & Accuracy [\%] \\ \hline \hline
    CK, start 40    & 60 & 74.60 & 60 & 75.36 \\
    Combined, start 40 & 60 & 74.68 & 60 & 75.37 \\
    Window, start 0 & 60 & 74.78 & - & - \\
    CK, start 30 & - & - & 80 & 73.66  \\ \hline
    PruneTrain (\cite{prunetrain})           & 50 & 73.0 & - & - \\
    Dyn Sparse (\cite{mostafa2019parameter}) & - & - & 80 & 73.3 \\ 
    Dyn Sparse (kernel granularity) & - & - & 80 & 72.6 \\\hline
    \end{tabular}
    \vspace{5pt}
    \caption{Main results, top-1 accuracy on ResNet-50 v1 after 90 and 100 epochs. Comparison with related work. PruneTrain sparsity is not explicitly stated, so we estimate their sparsity level from their inference FLOPs saved (1-FLOPs). Also, our experiments were run with batch size 64.}
    \vspace{-10pt}
    \label{tab:main_resnet_v1}
\end{table}

\begin{figure}[t]
    \centering
    \vspace{0pt}
    \includegraphics[width= .95 \textwidth]{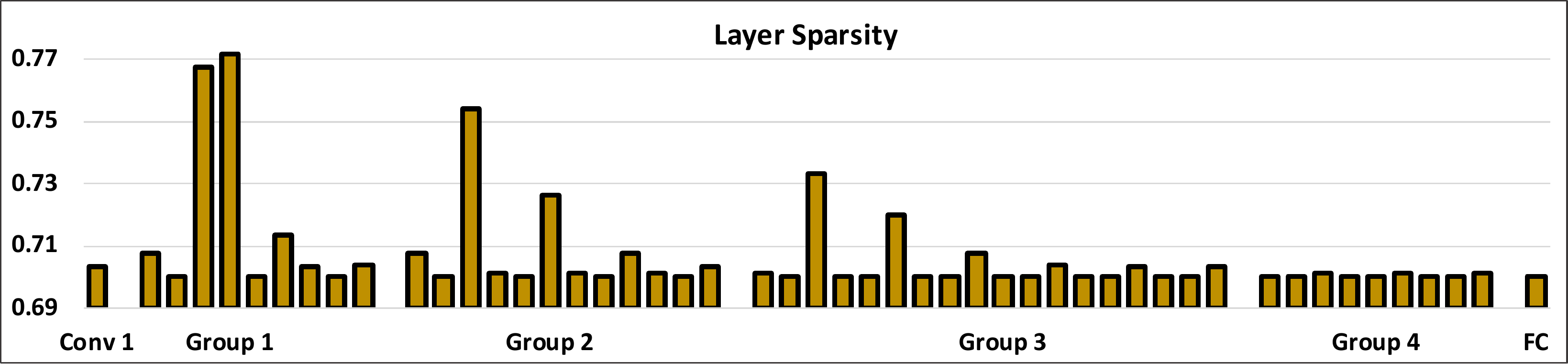}
    \vspace{-5pt}
    \caption{ResNet-50 v1.5 layer sparsity for CK, start 30, targetting 70\% sparsity.}
    \vspace{-5pt}
    \label{fig:Layer_sp}
\end{figure}

 Table~\ref{tab:window_resnet50_imagenet_adversarial} demonstrates that our sparsity mechanism can have a minimal drop in adversarial robustness (approximately 1-1.5\%) compared to the dense baseline model, whereas other methods see more accuracy degradation \cite{wang2018adv}. 

The sparsity of each layer, depicted in Figure~\ref{fig:Layer_sp}, emphasizes that early layers tolerate sparsity better, as they have consistently higher sparsity in the last 1$\times$1 convolutional layer of each residual block. This may be due to their vicinity to the residual connection, which provides additional information to the layer. 

Table \ref{tab:batch256_table} shows the results for select experiments with batch 256, which reflects more typical ResNet training. 

\begin{table}[t]
    
    \vspace{0pt}
    \centering
    \begin{tabular}{|c|c|c|} \hline
    Model Sparsity (\%) & 60 & 70 \\ \hline \hline
    Dense (76.53) & - & - \\ \hline
    CK, start 20 & - & 74.72 (-1.81) \\
    CK, start 30 & - & \textbf{75.13 (-1.40)} \\ \hline
    Window, start 20 & - & 74.68 (-1.85) \\
    Window, start 30 & - & 74.87 (-1.66) \\
    Window, start 40 & - & 75.03 (-1.50) \\ 
    Window, start 0, era 0-30 & 73.95 (-2.58) & - \\ \hline
    Combined, start 20 & - & 74.01 (-2.52) \\
    Combined, start 30 & - & 74.44 (-2.09) \\
    Combined, start 40 & - & 74.62 (-1.91) \\ \hline
    \end{tabular}
    \vspace{5pt}
    \caption{Main results, top-1 accuracy on ResNet-50 v1 after 100 epochs, using batch size 256.}
    \vspace{0pt}
    \label{tab:batch256_table}
\end{table}

\begin{table}[!ht]
    \vspace{0pt}
    \centering
    \begin{tabular}{|c |c | c|c|c|c|c|c|} \hline
         Model & Sparsity & \(\epsilon= .05\)& \(\epsilon=.1\) & \(\epsilon = .15\) & \(\epsilon=.2\) & \(\epsilon=.25\) & \(\epsilon=.3\)  \\ \hline
         Dense & 0 & 41.71 & 30.47 & 25.24 & 22.39 & 20.69 & 19.58 \\
         Combined, start 40 & 0.6 & 40.42 & 29.02 & 23.67 & 20.73 & 18.93 & 17.73 \\
         Combined, start 40 & 0.7 & 40.03 & 28.55 & 23.13 & 20.13 & 18.32 & 17.10 \\
         Window, start 30 & 0.6 & 40.52 & 29.12 & 23.77 & 20.81 & 18.99 & 17.78 \\
         Window, start 30 & 0.7 & 39.73 & 28.32 & 22.91 & 19.88 & 18.02 & 16.77 \\
         CK, start 40 & 0.6 & 40.51 & 29.21 & 23.95 & 21.08 & 19.34 & 18.18 \\
         CK, start 30 & 0.7 & 39.83 & 28.44 & 23.09 & 20.11 & 18.31 & 17.12 \\ \hline
    \end{tabular}
    \vspace{5pt}
    \caption{Adversarial Robustness of ResNet-50 v1.5 on Imagenet}
     \vspace{0pt}
    \label{tab:window_resnet50_imagenet_adversarial}
\end{table}

\subsection{Discussion}

Overall, we notice that there is a tolerance for sparsity (up to 70\%), which yields around 1\% accuracy loss compared to the dense baseline. However, this loss can be compensated by dropping the learning rate and performing another 10 epochs of training, which provides a 0.7-0.9\% accuracy increase. With high levels of sparsity this extension is computationally cheap.

We observed the early stages of dense training are important for high accuracy, as longer periods of dense training consistently outperformed shorter ones.
Moreover, widening the pruning era slightly (10 epochs) improves the final convergence accuracy (by around 0.2\%).

We also observed that pushing the learning rate drop schedule to earlier epochs or aligning it with pruning era does not improve the final accuracy. However, pushing the last learning rate drop from epoch 90 to 80 can improve the accuracy by around 0.1\% (See Appendix Table~\ref{tab:drop80} and Table~\ref{tab:main_results100}).

We postulate that window pruning performs worse for ResNet v1.5 compared to ResNet v1 due to the strided nature of convolutions in ResNet v1.5.

\section{Related Work}
\label{sec:related}

To give a broad comparison stage, we extended Mostafa and Wang's \cite{mostafa2019parameter} table on alternative sparsity mechanisms in Table \ref{tab:sparsity_work} with respect to characteristics of their mechanisms: training/compression focus, regularization, the period in which pruning is applied, strictness of parameter budget, and pruning granularity.
We explain each of the columns below: 

\begin{enumerate}

    \item \textbf{Training Focus}: Trying to train while maintaining/increasing sparsity of the network. The opposite is \textbf{Compression Focus}, i.e., methods that only seek to provide a smaller network for inference.  

    \item \textbf{Regularization}: Applying a regularization value to the loss, in order to find and prune irrelevant weights, while others use magnitude-based pruning.

    \item \textbf{Pruning Era}: The period during training in which the pruning is applied.

    \item \textbf{Strictness of Parameter Budget Era wrt to Pruning}:
    A strict parameter budget is fixed to the size of the final sparse model. Mostafa and Wang \cite{mostafa2019parameter} have a strict budget throughout training. Our method is only strict after the pruning era. 
    Some networks do not have a strict parameter budget and only prune weights that appear to be irrelevant and without a sparsity target. 
    
    \item \textbf{Pruning Granularity}: The level of granularity within in the network at which values are pruned.  For example, at the kernel level we determine which values to prune by examining only the weights in the kernel \cite{mao2017exploring}. See Figure~\ref{fig:3x3convprune} for more information.

\end{enumerate}
\begin{table}
\vspace{0pt}
\centering
\footnotesize

\begin{tabular}{|c|c|c|c|c|c|}
    \hline
\thead{Method}                                                  & \thead{Train/\\ Cmprss \\ Focus} & \thead{Requires \\ Regular\\-ization}& \thead{Pruning \\ Era}  & \thead{Strict \\ Parameter \\ Budget \\ Era}  & \thead{Granularity\\of Sparsity}  
\\ \hline
 Window (This Work)                                     & T                                 & No                         & Beginning           	  & After Pruning                                 & non in Window                    	 \\
 CK/Combined (This Work)                             & T                                 & No                         & Middle      			  & After Pruning                                 & Kernel                            	\\
 Evolutionary \cite{mocanu2018scalable}                  & T                                 & No                         & Beginning        	  & Throughout                                    & non-structured                    	\\
 Zhu and Gupta \cite{topruneornot}                                            & T                                 & No                         & Throughout			  & After Pruning                                 & non-structured                    	\\
   Lottery \cite{frankle2018lottery}                      & T                                 & No                        & Throughout             & Throughout                                   & non-structured                        \\
  RNN Pruning \cite{narang2017exploring}                        & T                                 & No                         & Beginning  			  & None                                          & non-structured                    	\\
  NeST\cite{dai2017nest}                    & T                                 & No                         & Throughout			  & None                                          & non-structured                    	\\
  Variational Dropout \cite{molchanov2017variational} & T                                 & No                        & Throughout			  & None                                    & non-structured                    	\\
   \hline
 PruneTrain \cite{prunetrain}                                   & T                                 & Yes                        & Throughout 			  & After Pruning                                 & Layer/Channel                     	\\
 Dyn Sparse \cite{mostafa2019parameter}                     & T                                 & Yes                              & Throughout		  & Throughout                                    & non-/Kernel                       	\\
 DeepR \cite{DBLP:journals/corr/abs-1711-05136} & T                                 & Yes                        & Throughout			  & Throughout                                    & non-structured                    	\\
 \hline
 \hline
 Deep Comp \cite{han2015learning}             & C                                 & No                         & Throughout			  & -                                             & non-structured                    	\\
 \hline
 L1-Norm Channel \cite{li2016pruning}                 & C                                 &Yes                         &Throughout			  & -                                             & Channel                           	\\
Brain Damage \cite{lebedev2016fast}                        & C                                 & Yes                        & End       			  & -                                             & non-structured                    	\\
 Sparsity Gran \cite{mao2017exploring}                   & C                                 & Yes                         & Throughout  		  & -                                              & non-structured                   	\\
SSL \cite{wen2016learning}                                      & C                                 & Yes                        & Throughout			  & -                                             & Channel/Kernel/Layer              	\\
 ThiNet \cite{luo2017thinet}                                    & C                                 & Yes                        & End       			  & -                                             & Channel                           	\\
 LASSO-regression \cite{he2017channel}                     & C                                 & Yes                        & End       			  & -                                             & Channel                           	\\
 Slimming \cite{liu2017learning}                        & C                                 & Yes                        & Throughout			  & -                                             & Channel                           	\\
 SSS \cite{huang2018data}                & C                                 & Yes                        & Throughout			  & -                                             & Layer                             	\\
 PFA \cite{suau2018principal}                                   & C                                 & Yes                        & Throughout			  & -                                             & Channel                           	\\ 
 \hline

\end{tabular}
\vspace{5pt}
\caption{Comparison of Training Methods that yield sparse networks } \label{tab:sparsity_work}
\vspace{-15pt}

\end{table}

We chose these concepts because their characteristics can enable faster and lower-energy training. 
A strict parameter budget allows the hardware mapping to plan for a fixed number of multiply-accumulate operations \cite{EIE}. 
Moreover, it allows a lower, fixed amount of physical memory to be allocated to an accelerator \cite{sohoni2019low,mostafa2019parameter,golub2018deep}.
The granularity of the sparsity mechanism indicates how easy it is to adapt the mechanism to an existing hardware. 
The coarser the granularity, the more adaptable it is to existing hardware~\cite{mao2017exploring}. 
Regularization, although useful in forcing the network to learn prunable weights, adds more non-linearity (and thus irregularity) to computation flow \cite{wu2018l1}. 
Pruning in the earlier epochs allows us to train with a compressed network for the majority of training. 

Mao et al. \cite{mao2017exploring} explores pruning on a range of granularities including window, kernel, and filter, and their effect on accuracy, using ImageNet on a number of CNN architectures, including ResNet-50, VGG, and GoogLeNet. 
They also qualitatively and quantitatively show that coarse-grain pruning, like kernel- or filter-level sparsity, is more energy-efficient due to fewer memory references.
Similarly, our work surveys sparsity at the window, kernel, and filter levels. We improve on Mao et al.'s work in two ways.
First, we show higher top-5 accuracy at higher sparsity levels on a complex benchmark, ImageNet on ResNet-50 (92.338\% at 40\% CK sparsity), and we also show high top-1 accuracy whereas Mao et al. only report top-5. 

Prunetrain \cite{prunetrain} explores a way to create sparse channels and even layers to speed up training with around a 1\% drop in accuracy. However, this requires a shift in the training mechanism, including a regularization term that could effect how the mechanism scales to large and distributed settings and that must be computed throughout training. 
The resulting network is only around 50\% sparse and the accuracy loss due to sparse training is high enough that a baseline network with same accuracy could result into same computational savings by just terminating training at much earlier stage/epoch. 

Gale et al. \cite{gale2019state} thoroughly characterize variational dropout \cite{molchanov2017variational}, L0-regularization \cite{louizos2017learning}, and Zhu and Gupta's \cite{topruneornot} magnitude-based pruning applied to Transformers and ImageNet on ResNet-50. Using larger batch size (1024) training they are able to achieve high accuracy (within 1\% decrease) compared to their baseline on ResNet-50 with variational dropout and magnitude-based pruning. However, L0-regularization was unable to produce sparsified networks without high loss in accuracy on ResNet-50, and the other two methods provide unstructured sparsity. In contrast, our work fixes the structure of the network early on in training, making our sparse training possible for hardware to accelerate.

In contrast to other pruning mechanisms, our proposed window, CK, and combined sparsity mechanisms have strict parameter budgets after the pruning era. The CK and combined schemes have channel-level and kernel-level pruning granularities.

\section{Conclusion and Future Work}
\label{sec:conclusion}

In this work, we introduced techniques to train CNNs with structured sparsity and studied the tradeoffs associated with various implementation options. We demonstrated on ResNet-50 with the full ImageNet dataset that the proposed sparse training method outperforms all related work and is comparable to a dense model in terms of convergence accuracy. 
We also observed that delaying the start of enforced, gradual pruning to at least epoch 20 was necessary to reach high convergence accuracy, highlighting the importance of the early epochs of dense training. 
Moreover, performing an additional 10 epochs of training provides substantial (around 1\%) accuracy gains of the final model.
In the future, we would like to study the tradeoffs of sparse training on low-precision networks.

\subsubsection*{Acknowledgments}
We thank Vitaliy Chiley, Sara O'Connell, Xin Wang, and Ilya Sharapov for their feedback on the manuscript.
This research was sponsored by NSF grants
CCF-1563113. Any opinions, findings and conclusions or recommendations expressed in this material are those of the authors and do not necessarily reflect the views of the National Science Foundation (NSF).

\newpage
\bibliographystyle{unsrt}
\bibliography{references}

\newpage

\appendix
\appendixpage

\section {Details of Pruning Algorithms}

Here we provide full descriptions of our other pruning methods and our general methodology sparse training.

\paragraph{Sparse Training Methodology}
Algorithm~\ref{alg:pruning_method} shows how we modify normal training in order to train sparsely.
\begin{algorithm}
    \caption{Pruning Algorithm}
    \label{alg:pruning_method}
    \begin{algorithmic}
    \STATE current\_iter = 0
    \WHILE{training}
        \IF{ current\_iter $>$ \textit{first epoch of pruning} and current\_iter $<$ \textit{last epoch of pruning} }
	    \STATE \textit{mask} = generate\_sparsity\_mask( $\bm{\theta}$, current\_iter, \textit{sparsity threshold} ) 
	
        \ENDIF
        \STATE $\bm{\theta}_{pruned} $ = \textit{mask} $\bigcap$ $\bm{\theta}$
        \STATE $\hat{\bm{y}}$ = forward\_pass( $\bm{\theta}_{pruned}$, $\bm{x}$ )
        \STATE $\bm{\theta}$ = weight\_update( $\bm{y}$, $\hat{\bm{y}}$, $\bm{\theta}_{pruned}$)
        \STATE current\_iter = current\_iter + 1
    \ENDWHILE
\end{algorithmic}
\end{algorithm}
\vspace{-10pt}

\paragraph{Window Pruning Methodology} Algorithm~\ref{alg:window_pruning} shows how we prune with window sparsity.

\begin{algorithm}
\caption{Window Pruning Algorithm}
\label{alg:window_pruning}
\begin{algorithmic}
\STATE generate\_window\_sparsity\_mask($\theta_{layer}$, sparsity\_threshold):
    \FOR{$\theta$ in $\theta_{layer}$}
        \FOR {all c in C}
            \FOR{all k in K}
                \STATE cutoff\_index = size($\theta_{c,k}$) $*$ sparsity\_threshold
                \STATE n = max(cutoff\_index, size($\theta_{c,k}$) $-$
                max\_non\_zero $-$ 1)
                \STATE cutoff\_value = $n^{th}$ largest value in $\theta_{c,k}$
                \FOR{all i,j in R,S}
                    \STATE mask$_{i,j,c,k}$ $=$ 1 if $\theta_{i,j,c,k}$ $>$ cutoff\_value, else 0
                \ENDFOR
            \ENDFOR
        \ENDFOR
    \ENDFOR
\end{algorithmic}
\end{algorithm}

\paragraph{Combined Pruning Methodology}

To combine Window and CK pruning, we introduce \textit{combined pruning}.
As shown by Algorithm~\ref{alg:combined_pruning}, in a given epoch we first apply Window Pruning to each 3$\times$3 convolutional layer at a fraction of the \textit{sparsity threshold} for that epoch. Then, we prune the remaining fraction of the \textit{sparsity threshold} with CK Pruning. The idea being that kernels that lose many of their parameters during window pruning can be fully pruned during the CK pruning phase. Our intuition is that first pruning parameters within a kernels guides the subsequent CK pruning towards the less important kernels. Thus, we pick out better kernels to prune. We also gain more structured sparsity but sacrifice the precision of window pruning.
\begin{algorithm}
\caption{combined Pruning Algorithm}
\label{alg:combined_pruning}
\begin{algorithmic}
\STATE generate\_combined\_sparsity\_mask($\theta_{layer}$, sparsity\_threshold):
    \FOR{$\theta$ in $\theta_{layer}$}
        \STATE window\_mask = generate\_ck\_sparsity\_mask($\theta$, sparsity\_threshold)
        \STATE ck\_mask = generate\_ck\_sparsity\_mask($\theta$, sparsity\_threshold)
        \STATE mask = window\_mask \AND ck\_mask
    \ENDFOR
\end{algorithmic}
\end{algorithm}
    
For completeness, we also tried another method of combining called \textit{inter-epoch pruning}, which involved splitting the \textit{pruning era} into CK pruning and window pruning phases. 
However, from our initial experiments we determined that combined pruning, performed better (though it was more computationally expensive) than inter-epoch pruning. With inter-epoch pruning we were only able to achieve 74.1\% top-1 accuracy with a \textit{first epoch of sparsity} of 40 and a \textit{final sparsity} of 40\% on ResNet-50v1.5 and Imagnet. The same setup trained with combined pruning achieved 74.9\% accuracy. Thus, we pursued combined pruning as our method to combine the two sparsification methods.

\section{Additional Details on Experimental Setup}

This section goes into more detail on the exact details of the models and dataset combinations we sued for experimentation.

\subsection{ResNet-50 on Tiny-Imagenet}

For this training domain, we trained using the Tiny-imagenet dataset~\cite{tinyimagenet} with resnet50~\cite{resnet50}. However, we changed the training mechanism in order to get validate our results. Each network we train for 40 epochs, with a batch size of 64. Additionally, we use the Adam optimizer to train with learning rate set to 0.001 and momentum set to 0.9. We also use weight decay set to 0.0001, and we anneal the learning rate to 0.0001 after 30 epochs of training in order to converge faster. We apply the same image transforms as on full Imagenet. 

We chose this optimization method because we felt that it achieved a good overall accuracy at a baseline level and represents the results in \cite{resnet50ontinyimagenet} in their vanilla model. We do not use the same preprocessing or image transforms in the report \cite{resnet50ontinyimagenet}.  Moreover, we wanted a quick way to estimate how our method would perform on full Imagenet. 

\subsection{ResNet-50 on Imagenet}

Here, we train each network for 90 epochs with a reduced batch size of 128 instead of 256 because 256 would not fit on a GPU in addition to our pruning layers. We found that changing the batch size to 128 but retaining all other hyperparameters as specified in \cite{resnet50} we were able to achieve the same benchmark 74.94\% accuracy as the paper. We train for 90 epochs with SGD with momentum set to 0.9 and weight decay is \num{1e-4}. We set the initial learning rate to be 0.1 and then anneal the learning rate to 0.01 at epoch 30 and then finally to 0.001 at epoch 60. 

For dataset transformations, we perform the same transformations as \footnote{\url{https://github.com/pytorch/examples/tree/master/imagenet}}. This means that during training we perform a random sized crop to size 224x224, randomly flip the image horizontally, and normalize the image. The batches are shuffled during training. For validation, we resize the image to 256 and then center crop to size 224x224 and then normalize. 

\subsection{ResNet-50v1.5 on Imagenet}

We train our ResNet-50 v1.5 \footnote{\url{https://ngc.nvidia.com/catalog/model-scripts/nvidia:resnet_50_v1_5_for_pytorch}} model for 90/100 epochs and use SGD with momentum (0.9) to optimize. The standard model says that learning rate should 0.1 for 256 batch size, but since that didn't fit in our GPUs with our sparsity mechanism, we used batch size 64 and linearly scaled the learning rate to be 0.05. We set the learning rate decay such that we multiply by 0.1 after 30, 60, and 90 epochs. We have weight decay set to \num{1e-4}.

\section{Miscellaneous Results}

\subsection{ResNet-50 on Tiny-Imagenet}

Our models actually perform better than the baseline with the following configurations: window pruning with 60\% sparsity as well as CK pruning with 20\%, 60\% and 80\% sparsity. The number of epochs required to reach the converge to the final accuracies is the same for CK and earlier for window at 40\% and 60\% sparsity.

\begin{figure}[t]
    \vspace{-10pt}
    \centering
    \begin{tikzpicture}
    \begin{axis}[cycle list name=tb, 
                    width=0.4\textwidth,
                    height=0.3\textwidth,
                    grid=both,
                    grid style={solid,gray!30!white},
                    axis lines=middle,
                    xlabel={Epoch},
                    ylabel={Accuracy},
                    xmin=20,
                    x label style={at={(axis description cs:0.5,-0.1)},anchor=north,font=\small},
                    y label style={at={(axis description cs:-0.1,.5)},rotate=90,anchor=south,font=\small},
                    legend pos=south east,
                    tick label style={font=\tiny},
                    legend style={font=\tiny}
                    ]
    \addplot table [x=, y=timagenet_resnet50_maxnonzero_2_20_batch64_s0_e10_f, col sep=comma]
    {timagenet_resnet50_1x1_prune_like_fc/Val_Prec_1_Avg.csv};
    \addlegendentry{20\%}
    \addplot table [x=, y=timagenet_resnet50_maxnonzero_2_40t_batch64_s0_e10_f, col sep=comma]
    {timagenet_resnet50_1x1_prune_like_fc/Val_Prec_1_Avg.csv};
    \addlegendentry{40\%}
    \addplot table [x=, y=timagenet_resnet50_maxnonzero_2_60t_batch64_s0_e10_f, col sep=comma]
    {timagenet_resnet50_1x1_prune_like_fc/Val_Prec_1_Avg.csv};
    \addlegendentry{60\%}
    \addplot table [x=, y=timagenet_resnet50_maxnonzero_2_80t_batch64_s0_f, col sep=comma]
    {timagenet_resnet50_1x1_prune_like_fc/Val_Prec_1_Avg.csv};
    \addlegendentry{80\%}
    \addplot table [x=, y=timagenet_resnet50_standard, col sep=comma]
    {timagenet_resnet50_standard/Val_Prec_1_Avg.csv};
    \addlegendentry{0\%}
    \end{axis}
    \end{tikzpicture}
\qquad
\begin{tikzpicture}
    \begin{axis}[cycle list name=tb, 
                    width=0.5\textwidth,
                    height=0.3\textwidth,
                    grid=both,
                    grid style={solid,gray!30!white},
                    axis lines=middle,
                    xlabel={Epoch},
                    ylabel={Accuracy},
                    x label style={at={(axis description cs:0.5,-0.1)},anchor=north,font=\small},
                    y label style={at={(axis description cs:-0.1,.5)},rotate=90,anchor=south,font=\small},
                    xmin=25,
                    xmax=31,
                    legend pos=south east,
                    tick label style={font=\tiny},
                    legend style={font=\tiny}
                    ]
    \addplot table [x=, y=timagenet_resnet50_kernel_prune_20_batch64, col sep=comma]
    {timagenet_resnet50_kernel_prune/Val_Prec_1_Avg.csv};
    \addlegendentry{20\%}
    \addplot table [x=, y=timagenet_resnet50_kernel_prune_40_batch64, col sep=comma]
    {timagenet_resnet50_kernel_prune/Val_Prec_1_Avg.csv};
    \addlegendentry{40\%}
    \addplot table [x=, y=timagenet_resnet50_kernel_prune_60_batch64, col sep=comma]
    {timagenet_resnet50_kernel_prune/Val_Prec_1_Avg.csv};
    \addlegendentry{60\%}
    \addplot table [x=, y=timagenet_resnet50_kernel_prune_80_batch64, col sep=comma]
    {timagenet_resnet50_kernel_prune/Val_Prec_1_Avg.csv};
    \addlegendentry{80\%}
    \addplot table [x=, y=timagenet_resnet50_standard, col sep=comma]
    {timagenet_resnet50_standard/Val_Prec_1_Avg.csv};
    \addlegendentry{0\%}
    \end{axis}
    \end{tikzpicture}

\caption{Convergence plots of ResNet-50 on Tiny-Imagenet with Window (left) and CK Pruning (right). At all sparsity levels we are the near or above the baseline. Though the differences between models is small, 60\% seems to performs the best and is a peak with respect to 40\% and 80\%.}
\label{fig:window_resnet50_timagenet}
\end{figure}
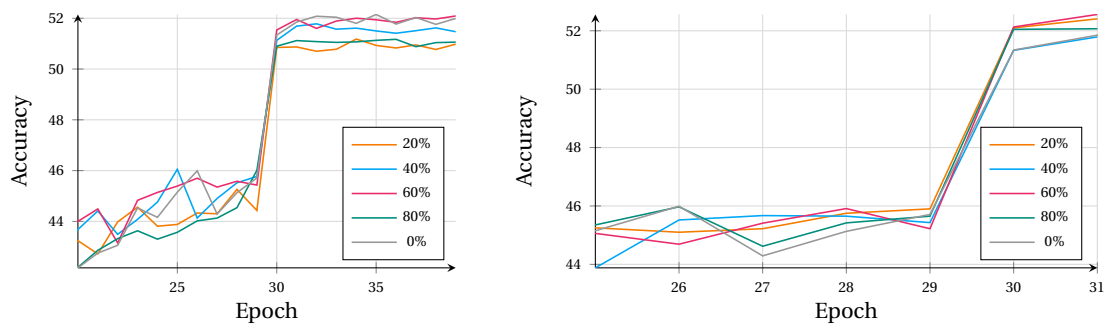

\begin{figure}[ht]
    \centering
    \begin{tikzpicture}
    \begin{axis}[cycle list name=tb, 
                    width=0.50\textwidth,
                    height=0.30\textwidth,
                    grid=both,
                    grid style={solid,gray!30!white},
                    axis lines=middle,
                    xlabel={Epoch},
                    ylabel={Accuracy},
                    x label style={at={(axis description cs:0.5,-0.1)},anchor=north,font=\small},
                    y label style={at={(axis description cs:-0.1,.5)},rotate=90,anchor=south,font=\small},
                    xmin=40,
                    xmax=100,
                    legend pos=south east,
                    tick label style={font=\tiny},
                    legend style={font=\tiny}
                    ]
    \addplot table [x=, y=resnet1_5_50_baseline-epoch-30-cksparse40-lr0-05out.out, col sep=comma]
    {ck_imagenet_resnet50v1_5/start30ck/Val_Prec_1_Avg.csv};
    \addlegendentry{40\%}
    \addplot table [x=, y=resnet1_5_50_baseline-epoch-30-cksparse60-lr0-05out.out, col sep=comma]
    {ck_imagenet_resnet50v1_5/start30ck/Val_Prec_1_Avg.csv};
    \addlegendentry{60\%}
    \addplot table [x=, y=resnet1_5_50_baseline-epoch-30-cksparse70-lr0-05out.out, col sep=comma]
    {ck_imagenet_resnet50v1_5/start30ck/Val_Prec_1_Avg.csv};
    \addlegendentry{70\%}
    \addplot table [x=, y=resnet1_5_50_baseline-epoch-30-cksparse80-lr0-05out.out, col sep=comma]
    {ck_imagenet_resnet50v1_5/start30ck/Val_Prec_1_Avg.csv};
    \addlegendentry{80\%}
    \end{axis}
    \end{tikzpicture}
\qquad 
    \begin{tikzpicture}
    \begin{axis}[cycle list name=tb,
                    width=0.45\textwidth,
                    height=0.30\textwidth,
                    grid=both,
                    grid style={solid,gray!30!white},
                    axis lines=middle,
                    xlabel={Epoch},
                    ylabel={Network Sparsity},
                    x label style={at={(axis description cs:0.5,-0.1)},anchor=north,font=\small},
                    y label style={at={(axis description cs:-0.1,.5)},rotate=90,anchor=south,font=\small},
                    xmax=100,
                    legend pos=south east,
                    tick label style={font=\tiny},
                    legend style={font=\tiny}
                    ]
    \addplot table [x=, y=resnet1_5_50_baseline-epoch-30-cksparse40-lr0-05out.out, col sep=comma]
    {ck_imagenet_resnet50v1_5/start30ck/Val_Sparsity.csv};
    \addlegendentry{40\%}
    \addplot table [x=, y=resnet1_5_50_baseline-epoch-30-cksparse60-lr0-05out.out, col sep=comma]
    {ck_imagenet_resnet50v1_5/start30ck/Val_Sparsity.csv};
    \addlegendentry{60\%}
    \addplot table [x=, y=resnet1_5_50_baseline-epoch-30-cksparse70-lr0-05out.out, col sep=comma]
    {ck_imagenet_resnet50v1_5/start30ck/Val_Sparsity.csv};
    \addlegendentry{70\%}
    \addplot table [x=, y=resnet1_5_50_baseline-epoch-30-cksparse80-lr0-05out.out, col sep=comma]
    {ck_imagenet_resnet50v1_5/start30ck/Val_Sparsity.csv};
    \addlegendentry{80\%}
    \end{axis}
    \end{tikzpicture}
    \vspace{-5pt}
\caption{Convergence and sparsity plots of ResNet-50v1.5 on Imagenet with CK Pruning, first epoch of sparsity = 30. As the amount of sparsity increases, the accuracy of the model seems to decrease semi-linearly. We can achieve up to 70\% while still being above 74\% sparsity.}
\label{fig:start30ck_resnet50v15_imagenet}
\end{figure}
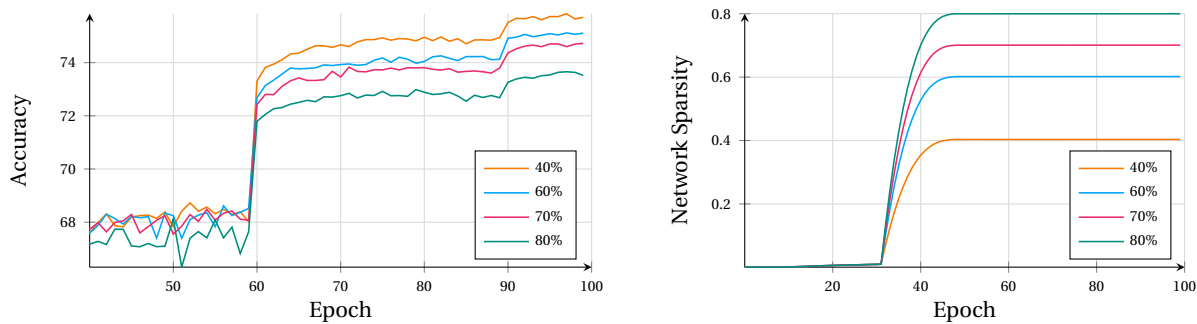

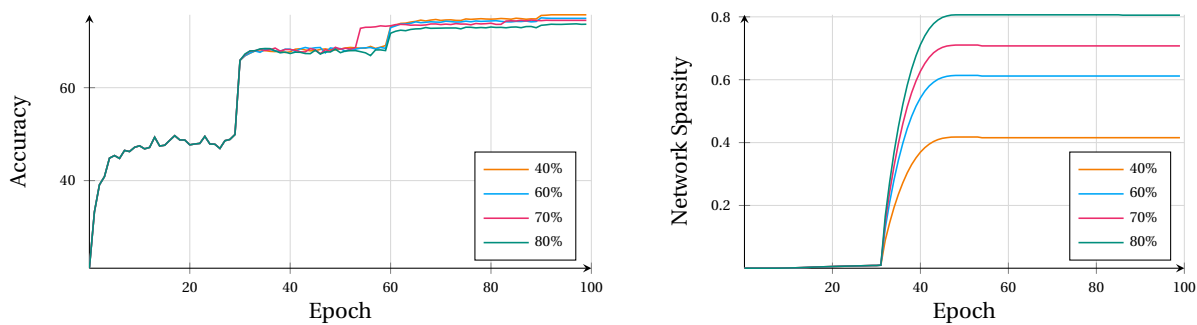
\begin{figure}[ht]
    \centering
    \begin{tikzpicture}
    \begin{axis}[cycle list name=tb, 
                    width=0.50\textwidth,
                    height=0.30\textwidth,
                    grid=both,
                    grid style={solid,gray!30!white},
                    axis lines=middle,
                    xlabel={Epoch},
                    ylabel={Accuracy},
                    x label style={at={(axis description cs:0.5,-0.1)},anchor=north,font=\small},
                    y label style={at={(axis description cs:-0.1,.5)},rotate=90,anchor=south,font=\small},
                    xmax=100,
                    legend pos=south east,
                    tick label style={font=\tiny},
                    legend style={font=\tiny}
                    ]
    \addplot table [x=, y=resnet1_5_50_baseline-epoch-30-cksparse40-lr0-05-intra-out.out, col sep=comma]
    {ck_imagenet_resnet50v1_5/start30intra/Val_Prec_1_Avg.csv};
    \addlegendentry{40\%}
    \addplot table [x=, y=resnet1_5_50_baseline-epoch-30-cksparse60-lr0-05-intra-out.out, col sep=comma]
    {ck_imagenet_resnet50v1_5/start30intra/Val_Prec_1_Avg.csv};
    \addlegendentry{60\%}
    \addplot table [x=, y=resnet1_5_50_baseline-epoch-30-cksparse70-lr0-05-intra-out.out, col sep=comma]
    {ck_imagenet_resnet50v1_5/start30intra/Val_Prec_1_Avg.csv};
    \addlegendentry{70\%}
    \addplot table [x=, y=resnet1_5_50_baseline-epoch-30-intra-combo-cksparse80-out.out, col sep=comma]
    {ck_imagenet_resnet50v1_5/start30intra/Val_Prec_1_Avg.csv};
    \addlegendentry{80\%}
    \end{axis}
    \end{tikzpicture}
\qquad 
    \begin{tikzpicture}
    \begin{axis}[cycle list name=tb,
                    width=0.45\textwidth,
                    height=0.30\textwidth,
                    grid=both,
                    grid style={solid,gray!30!white},
                    axis lines=middle,
                    xlabel={Epoch},
                    ylabel={Network Sparsity},
                    x label style={at={(axis description cs:0.5,-0.1)},anchor=north,font=\small},
                    y label style={at={(axis description cs:-0.1,.5)},rotate=90,anchor=south,font=\small},
                    xmax=100,
                    legend pos=south east,
                    tick label style={font=\tiny},
                    legend style={font=\tiny}
                    ]
    \addplot table [x=, y=resnet1_5_50_baseline-epoch-30-cksparse40-lr0-05-intra-out.out, col sep=comma]
    {ck_imagenet_resnet50v1_5/start30intra/Val_Sparsity.csv};
    \addlegendentry{40\%}
    \addplot table [x=, y=resnet1_5_50_baseline-epoch-30-cksparse60-lr0-05-intra-out.out, col sep=comma]
    {ck_imagenet_resnet50v1_5/start30intra/Val_Sparsity.csv};
    \addlegendentry{60\%}
    \addplot table [x=, y=resnet1_5_50_baseline-epoch-30-cksparse70-lr0-05-intra-out.out, col sep=comma]
    {ck_imagenet_resnet50v1_5/start30intra/Val_Sparsity.csv};
    \addlegendentry{70\%}
    \addplot table [x=, y=resnet1_5_50_baseline-epoch-30-intra-combo-cksparse80-out.out, col sep=comma]
    {ck_imagenet_resnet50v1_5/start30intra/Val_Sparsity.csv};
    \addlegendentry{80\%}
    \end{axis}
    \end{tikzpicture}
\caption{Convergence and sparsity plots of ResNet-50v1.5 on Imagenet with combined pruning, first epoch of sparsity = 30. The left plot shows training over the full 100 epochs instead of zooming in on the tail end of training. This allows us to observe the importance of the learning rate drops at epoch 30 and 60. The drop at epoch 90 does have a small increase, as well.}
\label{fig:start30intra_resnet50v15_imagenet}
\end{figure}

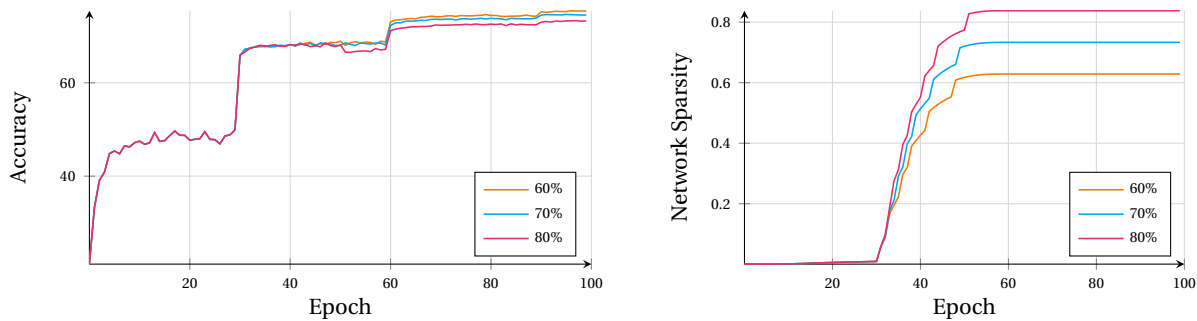
\begin{figure}[ht]
    \centering
    \begin{tikzpicture}
    \begin{axis}[cycle list name=tb, 
                    width=0.50\textwidth,
                    height=0.30\textwidth,
                    grid=both,
                    grid style={solid,gray!30!white},
                    axis lines=middle,
                    xlabel={Epoch},
                    ylabel={Accuracy},
                    x label style={at={(axis description cs:0.5,-0.1)},anchor=north,font=\small},
                    y label style={at={(axis description cs:-0.1,.5)},rotate=90,anchor=south,font=\small},
                    xmax=100,
                    legend pos=south east,
                    tick label style={font=\tiny},
                    legend style={font=\tiny}
                    ]

    \addplot table [x=, y=resnet1_5_50_baseline-epoch-30-window60-lr0-05out.out, col sep=comma]
    {ck_imagenet_resnet50v1_5/start30window/Val_Prec_1_Avg.csv};
    \addlegendentry{60\%}
    \addplot table [x=, y=resnet1_5_50_baseline-epoch-30-window70-lr0-05out.out, col sep=comma]
    {ck_imagenet_resnet50v1_5/start30window/Val_Prec_1_Avg.csv};
    \addlegendentry{70\%}
    \addplot table [x=, y=resnet1_5_50_baseline-epoch-30-window80-lr0-05out.out, col sep=comma]
    {ck_imagenet_resnet50v1_5/start30window/Val_Prec_1_Avg.csv};
    \addlegendentry{80\%}
    \end{axis}
    \end{tikzpicture}
\qquad 
    \begin{tikzpicture}
    \begin{axis}[cycle list name=tb,
                    width=0.45\textwidth,
                    height=0.30\textwidth,
                    grid=both,
                    grid style={solid,gray!30!white},
                    axis lines=middle,
                    xlabel={Epoch},
                    ylabel={Network Sparsity},
                    x label style={at={(axis description cs:0.5,-0.1)},anchor=north,font=\small},
                    y label style={at={(axis description cs:-0.1,.5)},rotate=90,anchor=south,font=\small},
                    xmax=100,
                    legend pos=south east,
                    tick label style={font=\tiny},
                    legend style={font=\tiny}
                    ]
    \addplot table [x=, y=resnet1_5_50_baseline-epoch-30-window60-lr0-05out.out, col sep=comma]
    {ck_imagenet_resnet50v1_5/start30window/Val_Sparsity.csv};
    \addlegendentry{60\%}
    \addplot table [x=, y=resnet1_5_50_baseline-epoch-30-window70-lr0-05out.out, col sep=comma]
    {ck_imagenet_resnet50v1_5/start30window/Val_Sparsity.csv};
    \addlegendentry{70\%}
    \addplot table [x=, y=resnet1_5_50_baseline-epoch-30-window80-lr0-05out.out, col sep=comma]
    {ck_imagenet_resnet50v1_5/start30window/Val_Sparsity.csv};
    \addlegendentry{80\%}
    \end{axis}
    \end{tikzpicture}
\caption{Convergence and sparsity plots of ResNet-50v1.5 on Imagenet with Window Pruning, first epoch of sparsity = 30. With window, likw with CK, the impact of the learning rate drops at epochs 30 and 60 is big. Note that the sparsity of the window is not as smooth as CK, showing that it is less uniformly sparse from epoch to epoch.}
\label{fig:start30window_resnet50v15_imagenet}
\end{figure}

\begin{figure}[ht]
    \centering
    \begin{tikzpicture}
    \begin{axis}[cycle list name=tb, 
                    width=0.50\textwidth,
                    height=0.30\textwidth,
                    grid=both,
                    grid style={solid,gray!30!white},
                    axis lines=middle,
                    xlabel={Epoch},
                    ylabel={Top-1 Accuracy},
                    x label style={at={(axis description cs:0.5,-0.1)},anchor=north,font=\small},
                    y label style={at={(axis description cs:-0.1,.5)},rotate=90,anchor=south,font=\small},
                    xmax=100,
                    xmin=30,
                    legend pos=outer north east,
                    tick label style={font=\tiny},
                    legend style={font=\tiny}
                    ]
    \addplot table [x=, y=resnet1_5_50_baseline-dense-lr0-05out.out, col sep=comma]
    {ck_imagenet_resnet50v1_5/sparse60/dense/Val_Prec_1_Avg.csv};
    \addlegendentry{Dense}
    \addplot table [x=, y=resnet1_5_50_baseline-epoch-0-window60-lr0-05out.out, col sep=comma]
    {ck_imagenet_resnet50v1_5/sparse60/start0/Val_Prec_1_Avg.csv};
    \addlegendentry{Window, Start 0}
    \addplot table [x=, y=resnet1_5_50_baseline-epoch-20-cksparse60-lr0-05out.out, col sep=comma]
    {ck_imagenet_resnet50v1_5/sparse60/start20/Val_Prec_1_Avg.csv};
    \addlegendentry{CK, Start 20}
    \addplot table [x=, y=resnet1_5_50_baseline-epoch-20-cksparse60-lr0-05-intra-out.out, col sep=comma]
    {ck_imagenet_resnet50v1_5/sparse60/start20/Val_Prec_1_Avg.csv};
    \addlegendentry{Combined, Start 20}
    \addplot table [x=, y=resnet1_5_50_baseline-epoch-30-cksparse60-lr0-05out.out, col sep=comma]
    {ck_imagenet_resnet50v1_5/sparse60/start30/Val_Prec_1_Avg.csv};
    \addlegendentry{CK, Start 30}
    \addplot table [x=, y=resnet1_5_50_baseline-epoch-30-cksparse60-lr0-05-intra-out.out, col sep=comma]
    {ck_imagenet_resnet50v1_5/sparse60/start30/Val_Prec_1_Avg.csv};
    \addlegendentry{Combined, Start 30}
    \addplot table [x=, y=resnet1_5_50_baseline-epoch-30-window60-lr0-05out.out, col sep=comma]
    {ck_imagenet_resnet50v1_5/sparse60/start30/Val_Prec_1_Avg.csv};
    \addlegendentry{Window, Start 30}
    \addplot table [x=, y=resnet1_5_50_baseline-epoch-40-cksparse60-lr0-05out.out, col sep=comma]
    {ck_imagenet_resnet50v1_5/sparse60/start40/Val_Prec_1_Avg.csv};
    \addlegendentry{CK, Start 40}
    \addplot table [x=, y=resnet1_5_50_baseline-lr05-epoch-40-intra-combo-cksparse60-out.out, col sep=comma]
    {ck_imagenet_resnet50v1_5/sparse60/start40/Val_Prec_1_Avg.csv};
    \addlegendentry{Combined, Start 40}
    \addplot table [x=, y=resnet1_5_50_baseline-era30-epoch-40-cksparse60-lr0-05out.out, col sep=comma]
    {ck_imagenet_resnet50v1_5/sparse60/start40_era30/Val_Prec_1_Avg.csv};
    \addlegendentry{CK, Start 40, Era 30}
    \addplot table [x=, y=resnet1_5_50_baseline-era30-lr05-epoch-40-intra-combo-cksparse60-out.out, col sep=comma]
    {ck_imagenet_resnet50v1_5/sparse60/start40_era30/Val_Prec_1_Avg.csv};
    \addlegendentry{Combined, Start 40, Era 30}

    \end{axis}
    \end{tikzpicture}
\qquad 
    \begin{tikzpicture}
    \begin{axis}[cycle list name=tb,
                    width=0.50\textwidth,
                    height=0.30\textwidth,
                    grid=both,
                    grid style={solid,gray!30!white},
                    axis lines=middle,
                    xlabel={Epoch},
                    ylabel={Top-5 Accuracy},
                    x label style={at={(axis description cs:0.5,-0.1)},anchor=north,font=\small},
                    y label style={at={(axis description cs:-0.1,.5)},rotate=90,anchor=south,font=\small},
                    xmax=100,
                    xmin=30,
                    legend pos=outer north east,
                    tick label style={font=\tiny},
                    legend style={font=\tiny}
                    ]
    \addplot table [x=, y=resnet1_5_50_baseline-dense-lr0-05out.out, col sep=comma]
    {ck_imagenet_resnet50v1_5/sparse60/dense/Val_Prec_5_Avg.csv};
    \addlegendentry{Dense}
    \addplot table [x=, y=resnet1_5_50_baseline-epoch-0-window60-lr0-05out.out, col sep=comma]
    {ck_imagenet_resnet50v1_5/sparse60/start0/Val_Prec_5_Avg.csv};
    \addlegendentry{Window, Start 0}
    \addplot table [x=, y=resnet1_5_50_baseline-epoch-20-cksparse60-lr0-05out.out, col sep=comma]
    {ck_imagenet_resnet50v1_5/sparse60/start20/Val_Prec_5_Avg.csv};
    \addlegendentry{CK, Start 20}
    \addplot table [x=, y=resnet1_5_50_baseline-epoch-20-cksparse60-lr0-05-intra-out.out, col sep=comma]
    {ck_imagenet_resnet50v1_5/sparse60/start20/Val_Prec_5_Avg.csv};
    \addlegendentry{Combined, Start 20}
    \addplot table [x=, y=resnet1_5_50_baseline-epoch-30-cksparse60-lr0-05out.out, col sep=comma]
    {ck_imagenet_resnet50v1_5/sparse60/start30/Val_Prec_5_Avg.csv};
    \addlegendentry{CK, Start 30}
    \addplot table [x=, y=resnet1_5_50_baseline-epoch-30-cksparse60-lr0-05-intra-out.out, col sep=comma]
    {ck_imagenet_resnet50v1_5/sparse60/start30/Val_Prec_5_Avg.csv};
    \addlegendentry{Combined, Start 30}
    \addplot table [x=, y=resnet1_5_50_baseline-epoch-30-window60-lr0-05out.out, col sep=comma]
    {ck_imagenet_resnet50v1_5/sparse60/start30/Val_Prec_5_Avg.csv};
    \addlegendentry{Window, Start 30}
    \addplot table [x=, y=resnet1_5_50_baseline-epoch-40-cksparse60-lr0-05out.out, col sep=comma]
    {ck_imagenet_resnet50v1_5/sparse60/start40/Val_Prec_5_Avg.csv};
    \addlegendentry{CK, Start 40}
    \addplot table [x=, y=resnet1_5_50_baseline-lr05-epoch-40-intra-combo-cksparse60-out.out, col sep=comma]
    {ck_imagenet_resnet50v1_5/sparse60/start40/Val_Prec_5_Avg.csv};
    \addlegendentry{Combined, Start 40}
    \addplot table [x=, y=resnet1_5_50_baseline-era30-epoch-40-cksparse60-lr0-05out.out, col sep=comma]
    {ck_imagenet_resnet50v1_5/sparse60/start40_era30/Val_Prec_5_Avg.csv};
    \addlegendentry{CK, Start 40, Era 30}
    \addplot table [x=, y=resnet1_5_50_baseline-era30-lr05-epoch-40-intra-combo-cksparse60-out.out, col sep=comma]
    {ck_imagenet_resnet50v1_5/sparse60/start40_era30/Val_Prec_5_Avg.csv};
    \addlegendentry{Combined, Start 40, Era 30}
    
    \end{axis}
    \end{tikzpicture}
\caption{Convergence plots of ResNet-50v1.5 on Imagenet at 60\%, top-1 and top-5 accuracy. For both top-1 an top-5, window pruning starting at 0 does not perform as well as the other methods. The rest are mostly clustered around 75\% top-1 accuracy and 92\% top-5 accuracy.}
\label{fig:sparse60_resnet50v15_imagenet_appendix}
\end{figure}
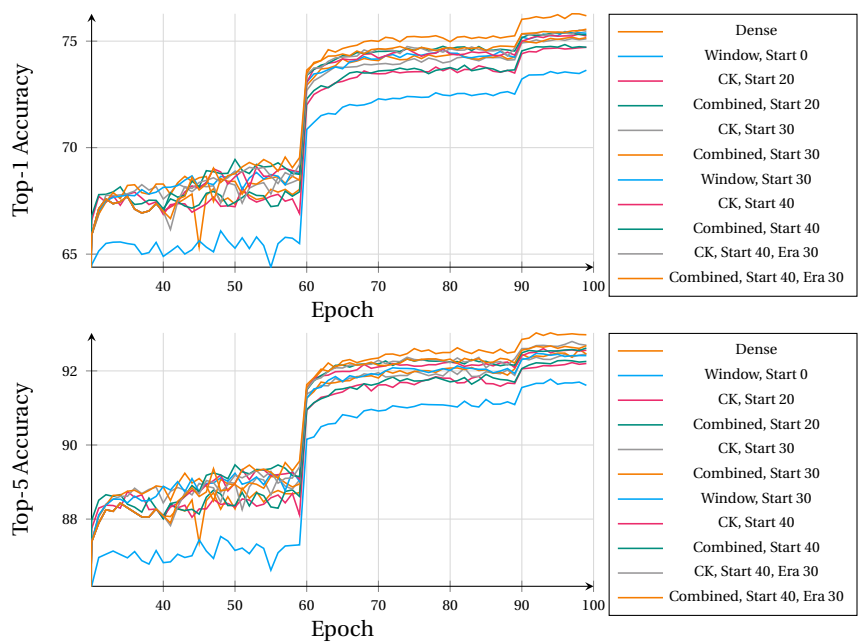

\begin{figure}[ht]
    \centering
    \begin{tikzpicture}
    \begin{axis}[cycle list name=tb, 
                    width=0.50\textwidth,
                    height=0.30\textwidth,
                    grid=both,
                    grid style={solid,gray!30!white},
                    axis lines=middle,
                    xlabel={Epoch},
                    ylabel={Accuracy},
                    x label style={at={(axis description cs:0.5,-0.1)},anchor=north,font=\small},
                    y label style={at={(axis description cs:-0.1,.5)},rotate=90,anchor=south,font=\small},
                    xmin=40,
                    xmax=100,
                    legend pos=south east,
                    tick label style={font=\tiny},
                    legend style={font=\tiny}
                    ]
    \addplot table [x=, y=resnet1_5_50_baseline-epoch-30-cksparse40-lr0-05out.out, col sep=comma]
    {ck_imagenet_resnet50v1_5/start30ck/Val_Prec_1_Avg.csv};
    \addlegendentry{40\%}
    \addplot table [x=, y=resnet1_5_50_baseline-epoch-30-cksparse60-lr0-05out.out, col sep=comma]
    {ck_imagenet_resnet50v1_5/start30ck/Val_Prec_1_Avg.csv};
    \addlegendentry{60\%}
    \addplot table [x=, y=resnet1_5_50_baseline-epoch-30-cksparse70-lr0-05out.out, col sep=comma]
    {ck_imagenet_resnet50v1_5/start30ck/Val_Prec_1_Avg.csv};
    \addlegendentry{70\%}
    \addplot table [x=, y=resnet1_5_50_baseline-epoch-30-cksparse80-lr0-05out.out, col sep=comma]
    {ck_imagenet_resnet50v1_5/start30ck/Val_Prec_1_Avg.csv};
    \addlegendentry{80\%}
    \end{axis}
    \end{tikzpicture}
\qquad 
\begin{tikzpicture}
    \begin{axis}[cycle list name=tb, 
                    width=0.45\textwidth,
                    height=0.30\textwidth,
                    grid=both,
                    grid style={solid,gray!30!white},
                    axis lines=middle,
                    xlabel={Epoch},
                    ylabel={Accuracy},
                    x label style={at={(axis description cs:0.5,-0.1)},anchor=north,font=\small},
                    y label style={at={(axis description cs:-0.1,.5)},rotate=90,anchor=south,font=\small},
                    xmin=40,
                    xmax=100,
                    legend pos=south east,
                    tick label style={font=\tiny},
                    legend style={font=\tiny}
                    ]
    \addplot table [x=, y=resnet1_5_50_baseline-epoch-30-cksparse40-lr0-05-intra-out.out, col sep=comma]
    {ck_imagenet_resnet50v1_5/start30intra/Val_Prec_1_Avg.csv};
    \addlegendentry{40\%}
    \addplot table [x=, y=resnet1_5_50_baseline-epoch-30-cksparse60-lr0-05-intra-out.out, col sep=comma]
    {ck_imagenet_resnet50v1_5/start30intra/Val_Prec_1_Avg.csv};
    \addlegendentry{60\%}
    \addplot table [x=, y=resnet1_5_50_baseline-epoch-30-cksparse70-lr0-05-intra-out.out, col sep=comma]
    {ck_imagenet_resnet50v1_5/start30intra/Val_Prec_1_Avg.csv};
    \addlegendentry{70\%}
    \addplot table [x=, y=resnet1_5_50_baseline-epoch-30-intra-combo-cksparse80-out.out, col sep=comma]
    {ck_imagenet_resnet50v1_5/start30intra/Val_Prec_1_Avg.csv};
    \addlegendentry{80\%}
    \end{axis}
    \end{tikzpicture}
\qquad 
    \begin{tikzpicture}
    \begin{axis}[cycle list name=tb, 
                    width=0.50\textwidth,
                    height=0.30\textwidth,
                    grid=both,
                    grid style={solid,gray!30!white},
                    axis lines=middle,
                    xlabel={Epoch},
                    ylabel={Accuracy},
                    x label style={at={(axis description cs:0.5,-0.1)},anchor=north,font=\small},
                    y label style={at={(axis description cs:-0.1,.5)},rotate=90,anchor=south,font=\small},
                    xmin=40,
                    xmax=100,
                    legend pos=south east,
                    tick label style={font=\tiny},
                    legend style={font=\tiny}
                    ]

    \addplot table [x=, y=resnet1_5_50_baseline-epoch-30-window60-lr0-05out.out, col sep=comma]
    {ck_imagenet_resnet50v1_5/start30window/Val_Prec_1_Avg.csv};
    \addlegendentry{60\%}
    \addplot table [x=, y=resnet1_5_50_baseline-epoch-30-window70-lr0-05out.out, col sep=comma]
    {ck_imagenet_resnet50v1_5/start30window/Val_Prec_1_Avg.csv};
    \addlegendentry{70\%}
    \addplot table [x=, y=resnet1_5_50_baseline-epoch-30-window80-lr0-05out.out, col sep=comma]
    {ck_imagenet_resnet50v1_5/start30window/Val_Prec_1_Avg.csv};
    \addlegendentry{80\%}
    \end{axis}
    \end{tikzpicture}
\qquad 
    \begin{tikzpicture}
    \begin{axis}[cycle list name=tb,
                    width=0.45\textwidth,
                    height=0.30\textwidth,
                    grid=both,
                    grid style={solid,gray!30!white},
                    axis lines=middle,
                    xlabel={Epoch},
                    ylabel={Network Sparsity},
                    x label style={at={(axis description cs:0.5,-0.1)},anchor=north,font=\small},
                    y label style={at={(axis description cs:-0.1,.5)},rotate=90,anchor=south,font=\small},
                    xmax=100,
                    legend pos=south east,
                    tick label style={font=\tiny},
                    legend style={font=\tiny}
                    ]
    \addplot table [x=, y=resnet1_5_50_baseline-epoch-30-window60-lr0-05out.out, col sep=comma]
    {ck_imagenet_resnet50v1_5/start30window/Val_Sparsity.csv};
    \addlegendentry{60\%}
    \addplot table [x=, y=resnet1_5_50_baseline-epoch-30-window70-lr0-05out.out, col sep=comma]
    {ck_imagenet_resnet50v1_5/start30window/Val_Sparsity.csv};
    \addlegendentry{70\%}
    \addplot table [x=, y=resnet1_5_50_baseline-epoch-30-window80-lr0-05out.out, col sep=comma]
    {ck_imagenet_resnet50v1_5/start30window/Val_Sparsity.csv};
    \addlegendentry{80\%}
    \end{axis}
    \end{tikzpicture}
    \vspace{-10pt}
\caption{Convergence plots with CK (top left), combined (top right), window (bottom left), and sparsity plot with window (bottom right). ResNet-50v1.5 on Imagenet with first epoch of sparsity = 30.}
\label{fig:start30_resnet50v15_imagenet}
\end{figure}
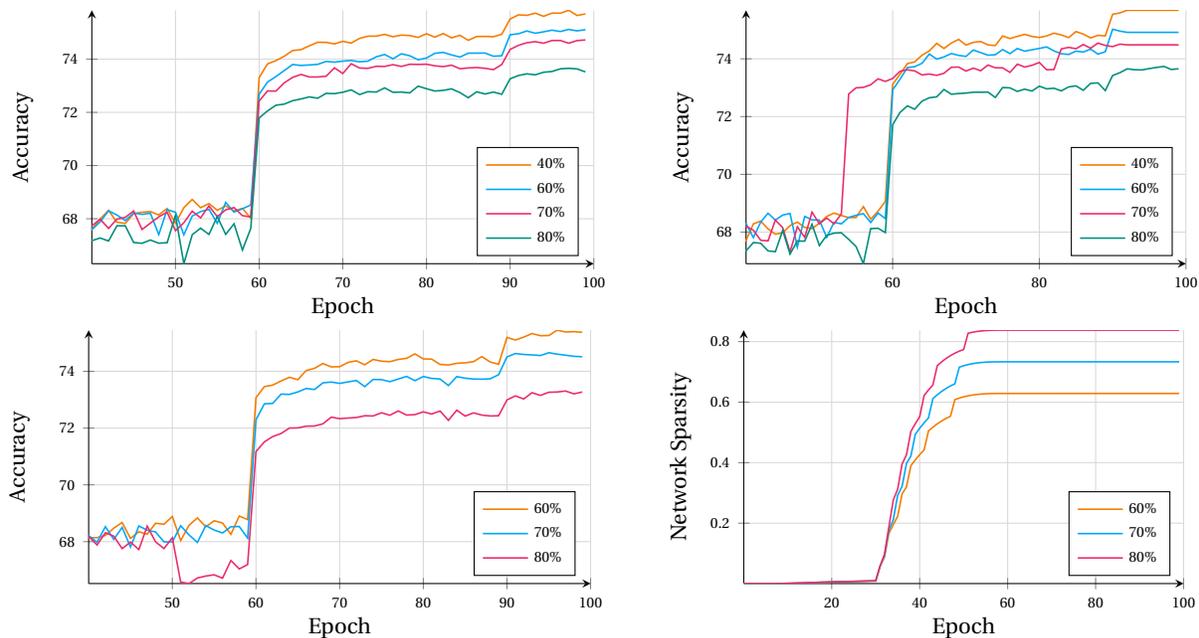

\begin{table}[ht]
    \footnotesize
    \centering
    \begin{tabular}{|c|c|c|c|c|} \hline
    Model Sparsity (\%) & 40 & 60 & 70 & 80 \\ \hline \hline
    Dense & 1.04 & - & - & - \\ \hline
    CK, start 40 & 0.89 & 0.73 & 0.71 & 0.83 \\
    CK, start 30 & 0.87 & 0.86 & 0.90 & - \\
    CK, start 20 & 0.80 & 0.87 & 0.96 & 0.30 \\ \hline
    Combined, start 40 & 0.98 & 0.64 & 0.72 & 0.79 \\ 
    Combined, start 30 &  0.89 & 0.75 & 0.61 & - \\ 
    Combined, start 20 & 0.99 & 1.10 & - & 0.75 \\ \hline
    Window, start 0 & - & 1.01 & 1.00 & 0.76 \\ 
    Window, start 30 & - & 1.17 & 0.77 & 0.67 \\ \hline
    CK, start 40, era 40-70    & - & 0.75 & 0.82 & - \\
    Combined, start 40, era 40-70 & - & 0.79 & 0.76 & - \\ \hline
    \end{tabular}
    \caption{Gain in top-1 accuracy on ResNet-50 v$1.5$ from epoch 90 to epoch 100. This table shows more explicitly that the benefit of the additional 10 epochs of training from epoch 90 to 100 is about 0.8-1.1\%.}
    \label{tab:main_results_compare}
\end{table}

\begin{table}[ht]
    \footnotesize
    \centering
    \begin{tabular}{|c|c|} \hline
    Experiment @ 60\% Sparsity & Accuracy (Improvement) [\%] \\ \hline \hline
    CK, start 40 & 75.39 (0.05) \\
    CK, start 30 & 75.06 (-0.06)\\
    CK, start 20 & 74.80 (0.09)\\ \hline
    Combined, start 40 & 75.46 (0.08) \\
    Combined, start 30 & 75.19 (0.03) \\
    Combined, start 20 & 74.88 (0.13) \\ \hline
    Window, start 0 & 73.89 (0.26) \\ \hline
    \end{tabular}
    \caption{Results when we changed final learning rate drop from epoch 90 to 80, top-1 accuracy on ResNet-50 v$1.5$ at epoch 100. This change does provide 0.1-0.2\% improvement is some cases, but the benefit is relatively small.}
    \label{tab:drop80}
\end{table}

\end{document}